\documentclass[runningheads]{llncs}

\usepackage{eccv}

\usepackage{hyperref}

\usepackage{orcidlink}

\usepackage{eccvabbrv}

\usepackage{graphicx}
\usepackage{booktabs}

\usepackage[accsupp]{axessibility}  

\usepackage{xcolor}
\usepackage{colortbl}
\usepackage{booktabs}
\usepackage{multirow}
\usepackage{macros}
\usepackage{enumitem}

\usepackage{tikz}
\usetikzlibrary{calc}
\usetikzlibrary{positioning}
\usepackage{pgfplots}
\usepackage{caption}
\usepackage{subcaption}
\pgfplotsset{compat=1.17}
\makeatletter
\renewcommand*{\thesubfigure}{%
  \ifnum\c@subfigure>26
    \@alph{\numexpr\c@subfigure-26\relax}\@alph{\numexpr\c@subfigure-26\relax}%
  \else
    \@alph{\c@subfigure}%
  \fi
}
\makeatother

\usepackage[table]{xcolor} 

\definecolor{highlight}{HTML}{33C9FF}
\definecolor{elobar}{HTML}{5AB643}

\newcommand{\seag}[2][highlight!20]{%
  \begingroup
  \setlength{\fboxsep}{2pt}
  \colorbox{#1}{#2}%
  \endgroup
}

\newcommand{\seagg}[2][highlight!40]{%
  \begingroup
  \setlength{\fboxsep}{2pt}
  \colorbox{#1}{#2}%
  \endgroup
}

\newcommand{\seaggg}[2][highlight!80]{%
  \begingroup
  \setlength{\fboxsep}{2pt}
  \colorbox{#1}{\textbf{#2}}%
  \endgroup
}

\allowdisplaybreaks[2]
\definecolor{original}{RGB}{0,0,255}      
\definecolor{composite}{RGB}{0,128,0}    
\definecolor{wasserstein}{RGB}{255,0,0}  
\definecolor{wassersteingan}{RGB}{0,0,0} 

\definecolor{first}{RGB}{220,50,47}    
\definecolor{second}{RGB}{255,140,0}   
\definecolor{third}{RGB}{255,193,7}    

\definecolor{arrow}{RGB}{142,211,251}    

\title{Drop-In Perceptual Optimization \\ for 3D Gaussian Splatting}

\author{
      Ezgi Özyılkan\inst{1,2}$^{*}$ \and
      Zhiqi Chen\inst{1}$^{*}$ \and
      Oren Rippel\inst{1} \and  \\
      Jona Ballé\inst{2} \and
      Kedar Tatwawadi\inst{1}$^{\dagger}$
  }
  \institute{
      Apple \\
      \and
      New York University Tandon School of Engineering \\
      \email{\{e\_ozyilkan, zhiqichen, oren.rippel\}@apple.com} \\
      \email{kedar.tatwawadi@gmail.com} \\
      \email{jona.balle@nyu.edu}
  }

\authorrunning{Özyılkan and Chen et al.}

\begin{document}

\maketitle

{\centering
\includegraphics[width=\textwidth, keepaspectratio]{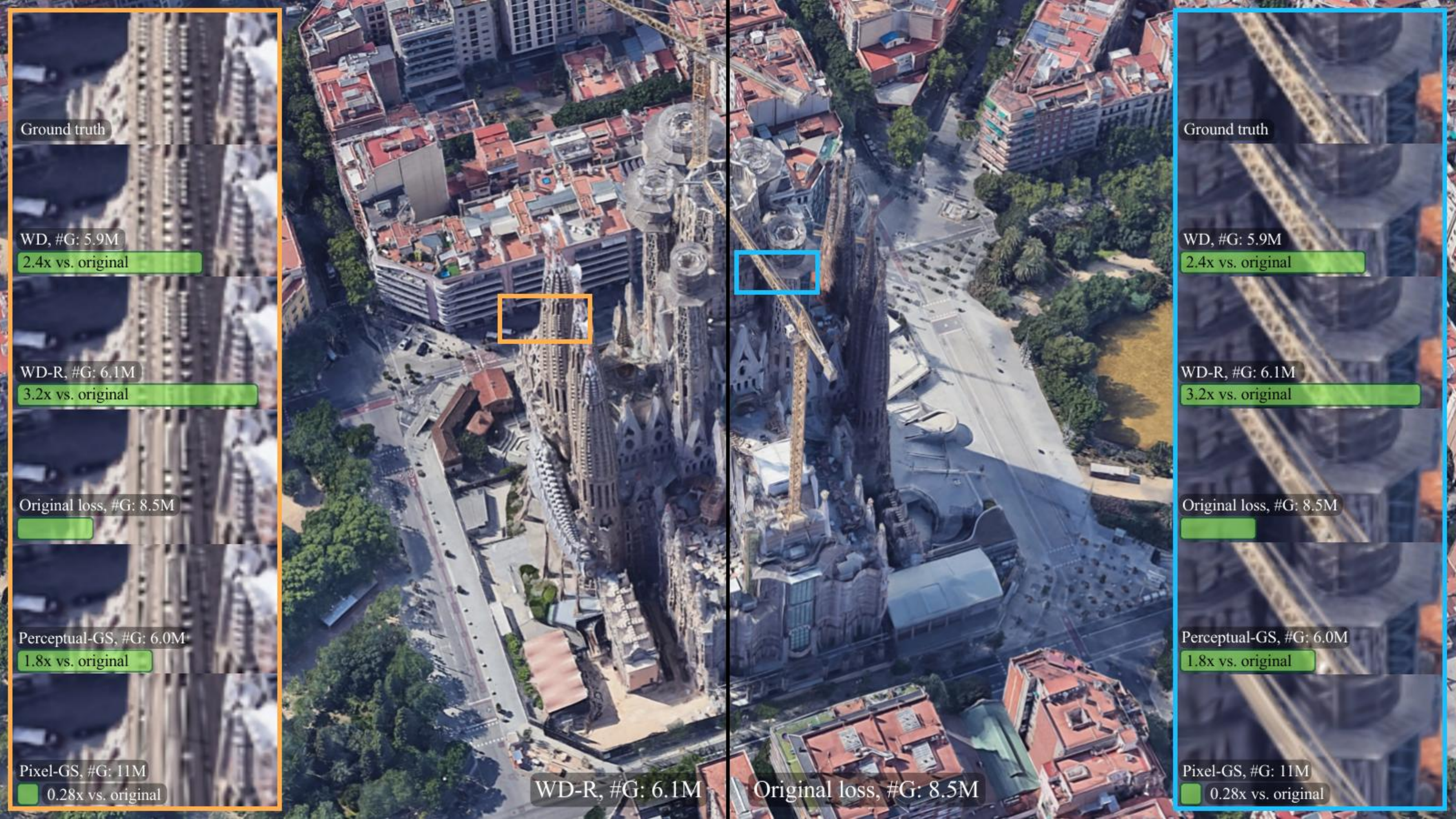}
\captionof{figure}{Novel view rendering using 3DGS on the large-scale \texttt{Barcelona} scene from the BungeeNeRF dataset~\cite{xiangli2022bungeenerf}. We compare the original 3DGS distortion loss $\mathrm{L1}+\mathrm{SSIM}$~\cite{kerbl3Dgaussians}, Pixel-GS~\cite{zhang2024pixel}, and the state-of-the-art in splat-efficient perceptual quality Perceptual-GS~\cite{zhou2025perceptual}, to the best-performing perceptual losses from our studies---Wasserstein Distortion~\cite{WD_orig} (WD) and a variant weighted with the original 3DGS loss, which we denote as WD-Regularized (\mbox{WD-R}). $\#\mathrm{G}$ denotes the splat count for each method, and \textcolor{elobar}{green bars} show the ratio of human raters preferring rendered image patches from the respective loss for this scene (\eg for every rater preferring the original loss, there are 2.4 and 3.2 raters preferring WD and WD-R, respectively).
\label{fig:teaser}}}

\def\thefootnote{*}\footnotetext{Equal contribution and corresponding authors. \\ This work was done during E.~Özyılkan's internship at Apple.
}
\def\thefootnote{$\dagger$}\footnotetext{Work done while at Apple.}
\def\thefootnote{\arabic{footnote}}

\begin{abstract}
Despite their output being ultimately consumed by human viewers, 3D Gaussian Splatting (3DGS) methods often rely on ad-hoc combinations of pixel-level losses, resulting in blurry renderings. To address this, we systematically explore perceptual optimization strategies for 3DGS by searching over a diverse set of distortion losses. We conduct the \mbox{first-of-its-kind} large-scale human subjective study on 3DGS, involving 39,320 pairwise ratings across several datasets and 3DGS frameworks. A regularized version of Wasserstein Distortion, which we call \mbox{\emph{WD-R}}, emerges as the clear winner, excelling at recovering fine textures without incurring a higher splat count. \mbox{WD-R} is preferred by raters more than $2.3\times$ over the original 3DGS loss, and $1.5\times$ over the current best method Perceptual-GS. \mbox{WD-R} also consistently achieves state-of-the-art LPIPS, DISTS, and FID scores across various datasets, and generalizes across recent frameworks, such as Mip-Splatting and Scaffold-GS, where replacing the original loss with WD-R consistently enhances perceptual quality within a similar resource budget (number of splats for Mip-Splatting, model size for Scaffold-GS), and leads to reconstructions being preferred by human raters $1.8\times$ and $3.6\times$, respectively. We also find that this carries over to the task of 3DGS scene compression, with $\approx 50\%$ bitrate savings for comparable perceptual metric performance.
\end{abstract}

\section{Introduction}
3D Gaussian Splatting (3DGS)~\cite{kerbl3Dgaussians} has emerged as a powerful technique for novel view synthesis, offering real-time and fully differentiable rendering. While the rendered output is intended for human viewers, most 3DGS methods predominantly rely on ad-hoc combinations of pixel-level losses, such as the original $\textrm{L1} + \textrm{SSIM}$ introduced by \cite{kerbl3Dgaussians}, which often lead to overly smooth or blurry textures. It is also common to use more recent distortion metrics such as LPIPS~\cite{zhang2018unreasonable} for evaluation, but their use for optimization has not been rigorously investigated. Importantly, the various perceptual optimization techniques have also not been validated through human preference studies. This lack of emphasis on human vision can lead to inefficient utilization of representational capacity, affecting rendering speed and real-time performance. This matters even more when a 3DGS representation must be compressed for storage or transmission, where every bit spent on imperceptible detail yields no perceptual benefit.

Several prior works study this gap, but mainly through partial or indirect strategies; \eg by enhancing edges~\cite{zhou2025perceptual}, or by manipulating the strategy of pruning and splitting Gaussians during training~\cite{gong2024eggs, zhang2024pixel}. In this work, we take a principled approach: we make the perceptual optimization loss itself the central design choice. Decoupling the modeling of human vision in the distortion loss from the underlying 3DGS algorithms---such as initialization, densification, and pruning---also simplifies the overall system design and enables broader generalization.

Building on this principle, in this work we take a systematic look at perceptual optimization for 3DGS, demonstrating that significant perceptual improvements can be achieved with \textit{improved loss formulation alone}, without relying on any heuristics or changes to the underlying 3DGS algorithms themselves:
\begin{enumerate}[itemsep=0pt, topsep=0pt, parsep=0pt]
    \item We study three families of distortion losses with the goal of maximizing perceptual quality for 3DGS: (i)~$\textrm{L1}+\textrm{SSIM}$ originally proposed in~\cite{kerbl3Dgaussians} and widely used across the literature, (ii)~a composite loss $\textrm{L1}+\textrm{L2}+\textrm{MS-SSIM}+\textrm{LPIPS}$ including common perceptual metrics~\cite{mentzer2020highfidelitygenerativeimagecompression, muckley_implicit}, and (iii)~Wasserstein Distortion (WD)~\cite{WD_Qiu, WD_orig}, a recently proposed distortion metric comparing local statistics in a deep feature space, which has shown promise in the learned compression literature~\cite{WD_orig}.
    \item To assess the perceptual quality achieved by these training objectives, we conduct a large-scale human subjective study comprising 39,320 pairwise ratings across 4 datasets and 3 distinct 3DGS frameworks---to our knowledge, the \mbox{first-of-its-kind} for 3DGS.
    \item Out of our analysis emerges a new loss as a top performer for 3DGS perceptual optimization. It is a regularized version of WD, which we refer to as \textit{\mbox{WD-R}}, that consistently outperforms all other losses. \mbox{WD-R} is preferred by raters more than $2.3\times$ over the original 3DGS loss and $1.5\times$ over Perceptual-GS, the current best perceptual 3DGS scheme. \mbox{WD-R} also achieves state-of-the-art scores on common perceptual metrics such as FID~\cite{FID_orig}, CMMD~\cite{CMMD_orig}, DISTS~\cite{Ding_2020}, and LPIPS~\cite{zhang2018unreasonable} across diverse datasets. We further show that \mbox{WD-R} generalizes to anti-aliasing and structured methods, as well as scene compression (\ie fitting 3DGS representations with a constraint on storage size rather than splat count), where it yields $\approx 50\%$ bitrate savings at comparable perceptual metric performance.
\end{enumerate}
Our contribution is realized through a simple yet effective modification---replacing the 2D distortion loss function used during optimization, without introducing any 3D-specific constraints or heuristics. As such, other extensions of the 3DGS framework (such as~\cite{zhang2024pixel, zhang2024fregs, yu2024mip, lu2024scaffold}) are largely orthogonal to and compatible with our work. Our project page can be found at \url{https://apple.github.io/ml-perceptual-3dgs}.


\section{Background}
\subsection{Related work}
Several approaches have sought to improve the visual quality of rendered 3DGS views. Mip-Splatting~\cite{yu2024mip} and Analytic-Splatting~\cite{liang2024analytic} reduce aliasing artifacts with improved filtering. Other works focus on optimization and primitive allocation, using frequency regularization (FreGS~\cite{zhang2024fregs}) or edge- or gradient-based heuristics to guide densification toward visually important regions (\eg Pixel-GS~\cite{zhang2024pixel}, Perceptual-GS~\cite{zhou2025perceptual}, and EGGS~\cite{gong2024eggs}). A third line of work explores more structured Gaussian representations for improved efficiency and compactness, such as anchor-based Scaffold-GS~\cite{lu2024scaffold}.

Perceptual realism is commonly pursued through adversarial training~\cite{mentzer2020highfidelitygenerativeimagecompression, muckley_implicit}, and GAN-based methods have been explored for 3DGS representations~\cite{kirschstein2024gghead,barthel2024gaussian, barthel2025cgsgan} in specialized domains such as faces or avatars. However, these approaches typically rely on domain-specific priors and lack the ``plug-and-play'' flexibility needed for general-purpose 3DGS optimization. As a result, perceptual optimization for arbitrary scenes remains underexplored.

The above-mentioned advances often apply modifications or constraints to the model, the optimization procedure, and the loss at the same time, in order to achieve better visual quality. In contrast, our approach cleanly separates perceptual modeling from algorithmic design, delegating it only to the distortion in the loss function. We additionally corroborate these gains using a large-scale human preference study, which, to the best of our knowledge, has never been conducted in any prior 3DGS work.

\begin{figure*}[t]
    \centering
    \includegraphics[width=\linewidth]{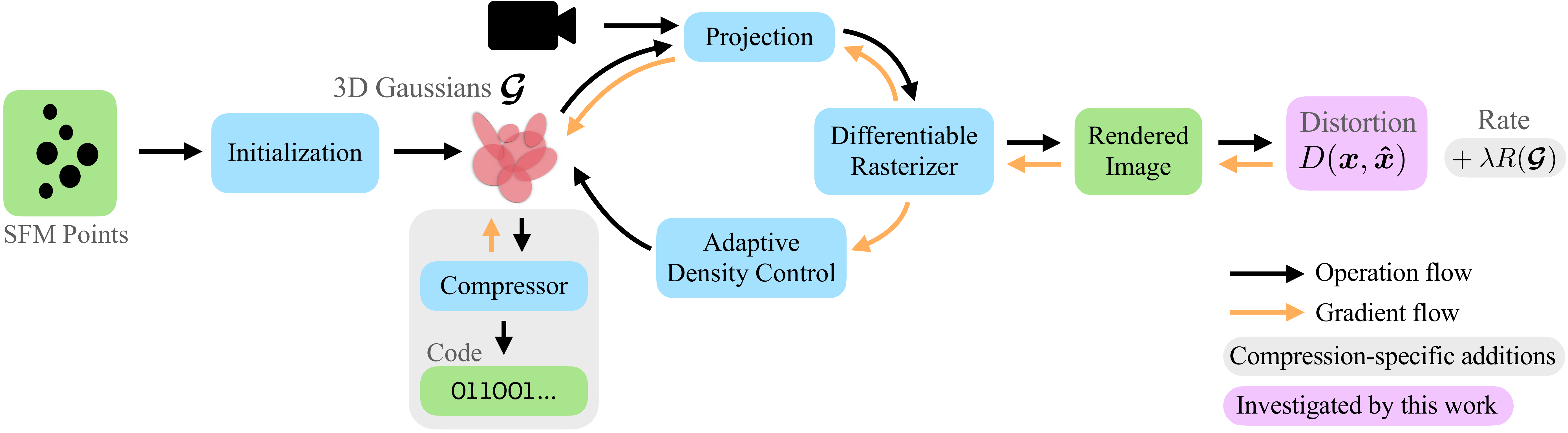}
    \caption{3DGS representation and compression frameworks optimized using~\eqref{eq:2D_distortion} and~\eqref{eq:RD}, respectively, incorporating the perceptual losses discussed in~\cref{subsec:perceptual_losses}.} \label{fig:framework}
\end{figure*}

\subsection{3DGS optimization procedure}
3DGS represents scenes as a collection of 3D Gaussian primitives that are differentiably rendered to produce 2D images. Each Gaussian primitive $G_i$ is parameterized by $G_i(\bmu_i, \bSigma_i, \bc_i, \alpha_i)$ where $\boldsymbol{\mu}_i$ is the center, $\boldsymbol{\Sigma}_i$ is the covariance matrix, $\boldsymbol{c}_i$ is color, and $\alpha_i$ is opacity~\cite{kerbl3Dgaussians}. During rendering, 3D Gaussians are projected to two dimensions and alpha-blended in depth order to produce the final pixel values of the reconstructed 2D view. As all the rendering operations are differentiable, this allows gradient-based optimization of the Gaussian parameters using just a distortion loss imposed on the rendered views.

The number of Gaussians is modulated dynamically during training via \emph{adaptive densification}, which introduces new primitives based on gradient magnitudes and removes low-opacity ones through pruning. These decisions are driven by gradients from the distortion loss, meaning the optimization objective influences not only parameter updates but also the evolution of the representation itself. Lower thresholds allow increased growth and more complex detail and texture reconstruction at the cost of increased runtime and memory, while higher thresholds favor compact, faster real-time renderings.

The overall optimization procedure is illustrated in~\cref{fig:framework}. During training, rendered views $\bhx$ are compared against ground-truth images using one of our proposed distortion losses (described in~\cref{subsec:perceptual_losses}), and the resulting gradients guide the optimization of Gaussian parameters and densification. During evaluation, all experiments measure performance on the \emph{novel view synthesis} task, where reconstructed images are compared against unseen test views.

We consider two settings with different optimization objectives: (i)~evaluate perceptual training objectives in the standard 3DGS representation framework~\cite{kerbl3Dgaussians}, and (ii)~examine whether the gains obtained by the best-performing perceptual strategy generalize to alternative 3DGS frameworks, including anti-aliasing rendering methods~\cite{yu2024mip} and structured Gaussian representations~\cite{lu2024scaffold}, as well as to a scene compression framework~\cite{liu2024compgs}.

\paragraph{\textbf{Representation}}
For the representation task, training minimizes a distortion loss between the ground-truth image $\bx$ and the rendered image $\bhx$:
\begin{equation}
\min_{\bmcG} \gamma \, D(\bx, \bhx),\label{eq:2D_distortion}
\end{equation} where $\bmcG$ is the collection of Gaussian parameters, $D(\cdot)$ denotes distortion computed on the rendered 2D image space, and $\gamma$ is a scalar weighting factor. Because adaptive densification relies on gradient magnitude, the scale of the loss indirectly affects the trade-off between reconstruction quality and the number of Gaussians in the 3DGS representation. To ensure a fair comparison under similar representation budgets, we allow $\gamma$ to vary across datasets while keeping all other hyperparameters fixed, in order to align the resulting Gaussian counts across methods without modifying the underlying framework.

\paragraph{\textbf{Variable-rate compression}}
In the task of scene compression~\cite{liu2024compgs, wang2024contextgs}, distortion and model size are balanced through a rate--distortion objective, as in learned image compression~\cite{balle2016opt,rippel17, balle2018variational}:
\begin{equation} \label{eq:RD}
    \min_{\bmcG, \boldsymbol{\theta}} D_{\boldsymbol{\theta}}(\bx, \bhx) + \lambda R_{\boldsymbol{\theta}}(\bmcG),
\end{equation} where $R_{\boldsymbol{\theta}}(\bmcG)$ estimates the storage cost of Gaussian parameters $\bmcG$, given the compression network parameters $\boldsymbol{\theta}$. Concretely, we use the common rate term used in neural compression $R_{\boldsymbol{\theta}}(\bmcG) = \mathbb{E}[-\log_{2} p_{\hat{\mathbf{y}}}(\hat{\mathbf{y}})]$, the Shannon cross-entropy between the quantized latent encoding $\hat{\mathbf{y}}$ of $\bmcG$ and a learned parametric entropy model $p_{\hat{\mathbf{y}}}$~\cite{balle2016opt}. The trade-off between reconstruction quality and storage cost is controlled by $\lambda$. The storage size of a 3DGS scene is not determined solely by the number of Gaussians, but also depends on parameter redundancy and the probabilistic modeling used for entropy coding.

\section{Methodology}
\subsection{Alternatives for the distortion loss $D$}
\label{subsec:perceptual_losses}
The original 3DGS formulation \cite{kerbl3Dgaussians} optimizes a linear combination of $\textrm{L1}$ and $\textrm{SSIM}$~\cite{SSIM}; this is the canonical choice for 3DGS loss~\cite{zhang2024pixel,zhou2025perceptual,yu2024mip,lu2024scaffold,liu2024compgs}, and we refer to it as the \emph{original loss}. While computationally efficient, these pixel-level losses often poorly correlate with human perception~\cite{MS-SSIM}. To explore alternatives better aligned with perceptual quality, we evaluate three additional loss formulations.

\paragraph{\textbf{Composite loss}} Since the introduction of SSIM as a model of human contrast perception~\cite{SSIM}, the field of human perceptual modeling has made significant advances. As a stronger baseline, we therefore consider a generalization of the \emph{original loss}, employing a weighted sum of several more recent and commonly used distortion metrics:
\begin{equation}
\resizebox{.65\hsize}{!}{$\mathcal{L}_{\textrm{composite}} = \omega_{1}\mathcal{L}_{\textrm{L1}} + \omega_{2}\mathcal{L}_{\textrm{L2}} + \omega_{3}\mathcal{L}_{\textrm{MS-SSIM}} + \omega_{4}\mathcal{L}_{\textrm{LPIPS}}$},
\label{eq:composite_loss}
\end{equation}
where $\mathcal{L}_{\textrm{L1}}$ and $\mathcal{L}_{\textrm{L2}}$ are per-pixel L1 and L2 norms, $\mathcal{L}_{\textrm{MS-SSIM}}$ is the multi-scale structural similarity loss~\cite{MS-SSIM}, $\mathcal{L}_{\textrm{LPIPS}}$~\cite{zhang2018unreasonable} measures perceptual distance using features extracted from a pretrained deep neural network, and $\omega_{1}$--$\omega_{4}$ control the relative importance of each term.

For LPIPS, we train with LPIPS-AlexNet while evaluating with LPIPS-VGG to avoid overfitting to the training metric and remain consistent with common evaluation practices in the 3DGS literature~\cite{kerbl3Dgaussians, zhang2024pixel, zhou2025perceptual}. The weights of the composite loss components are selected through ablation experiments (\cref{appendix_sec:ablation_composite}), and the configuration $\omega_1=0.05$, $\omega_2=0.30$, $\omega_3=0.60$, $\omega_4=0.10$ is used for all the experiments for its favorable trade-off between reconstruction quality and representation efficiency.

\paragraph{\textbf{Wasserstein distortion (WD)}}
LPIPS~\cite{zhang2018unreasonable} was the first metric to demonstrate that deep feature representations can approximate aspects of human visual perception~\cite{yamins-2016-using-goal-driven}. However, it is a \emph{pointwise} metric, which does not account for the way humans perceive \emph{texture}. \cref{fig:wd_intuition} illustrates this limitation in 1D: two textures with a $180^\circ$ phase shift exhibit large pointwise differences despite appearing visually similar. WD~\cite{WD_Qiu} is a feature-space agnostic distortion metric that compares spatially local estimates of statistics, allowing textures that differ substantially pointwise to be considered perceptually similar when their local statistics match. This formulation is motivated by models of the human visual system suggesting that peripheral vision encodes images using summary statistics over local pooling regions, rather than precise pointwise pixel values~\cite{freeman2011metamers, rosenholtz2011visual}. In this work, we use a version of WD where the spatially local statistics are computed in the VGG feature space~\cite{WD_orig}. For each feature map and spatial location, WD computes the RMSE between the local mean $\mu$ and standard deviation $\nu$:
\begin{equation}
    d_{\textrm{WD}} = \sqrt{(\mu - \hat{\mu})^2 + (\nu - \hat{\nu})^2}, \label{eq:WD}
\end{equation}
where $\hat{\phantom{x}}$ indicates the local statistics computed on the reconstructed image rather than the ground truth. These local differences are then aggregated over feature maps and pixel locations.
\begin{figure}[t]
    \definecolor{color_x}{HTML}{007EAA}
    \definecolor{color_x_hat}{HTML}{942192}
    \centering
    \begin{minipage}{0.58\linewidth}
        \centering
        \includegraphics[width=\linewidth]{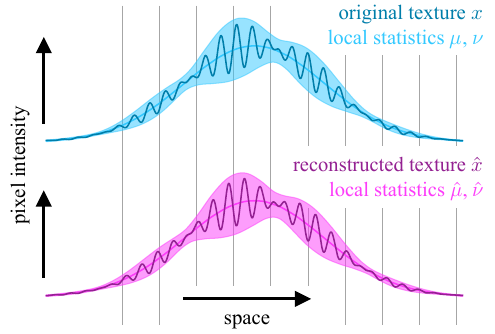}
    \end{minipage}
    \hfill
    \begin{minipage}{0.41\linewidth}
        \caption{1D sketch of visual textures with large pointwise difference but small Wasserstein distortion (WD) (see~\cref{eq:WD}). Both \textcolor{color_x}{original} and \textcolor{color_x_hat}{reconstructed} textures look the same at first glance, and share nearly the same local mean and standard deviation. However, they are very different in terms of pointwise differences. In contrast to all other metrics we compare here, WD evaluates differences in terms of local statistics (\eg the first and second moments, $\bmu$ and $\bnu$).} \label{fig:wd_intuition}
    \end{minipage}
\end{figure}

The area over which the local statistics are computed is determined by a pooling kernel of size $\sigma$, which can vary across locations. Larger $\sigma$-values capture texture realism by aggregating information over broader regions, while $\sigma \rightarrow 0$ converges to a pointwise distance. This can be used to make the metric more permissive to texture resampling in certain regions such as ones that are less visually salient. In this work, we use a constant $\sigma=4$, selected empirically (\cref{appendix_sec:ablation_wd_sigma}) at the resolutions we benchmarked; we therefore interpret it as a strong setting for our evaluation setup rather than a universal optimum, since the optimal $\sigma$ may depend on input image resolution and other parameters such as viewing distance.

\paragraph{\textbf{WD-regularized (\mbox{WD-R})}} In practice, we found that including a small amount of pixel-level fidelity loss along with WD helps to avoid artifacts in regions where the number of splats is constrained, and the number of training views is small (see~\cref{subsec:3dgs_representation}). We therefore introduce \emph{WD-Regularized} (\mbox{WD-R}), which augments WD with a lightly weighted version of the original loss. In all the experiments, WD still remains the primary objective, and incorporating the original loss serves only as modest regularization. Concretely, we optimize $\mathcal{L}_{\text{WD}} = \gamma\, d_{\text{WD}}$ for the WD setting and $\mathcal{L}_{\text{WD-R}} = \gamma\, (d_{\text{WD}} + \beta\, \mathcal{L}_{\text{orig}})$ for WD-R, where $\mathcal{L}_{\text{orig}}$ is the original 3DGS pixel-level loss~\cite{kerbl3Dgaussians}, $\gamma$ is a dataset-specific global scale (\cref{eq:2D_distortion}) used to align Gaussian counts across methods, and $\beta$ is chosen so that $\mathcal{L}_{\text{orig}}$ acts as a mild pixel fidelity regularizer while WD remains the dominant perceptual term. More details are discussed in~\cref{appendix_subsec:loss_hyperparam}.

\subsection{Training and evaluation setup} \label{sec:evaluation}
\paragraph{\textbf{Implementation details}}
We follow the experimental setup of prior work~\cite{kerbl3Dgaussians} and train all models for 30k iterations, adopting the published configurations of the respective baselines to ensure faithful reproduction. For WD-based training, similar to the warm-up strategies used in perceptual and adversarial training~\cite{isola2017image, ledig2017photo}, we first optimize the model using the standard 3DGS loss for 3k--5k iterations before introducing the perceptual objective.

For the representation task, we compare against \emph{Pixel-GS}~\cite{zhang2024pixel} and \emph{Perceptual-GS}~\cite{zhou2025perceptual}, two recent methods that improve perceptual quality by modifying gradient accumulation and densification strategies. These methods are particularly relevant baselines as they aim to improve perceptual sharpness while explicitly controlling Gaussian growth in order to maintain fewer splats---a constraint not always prioritized in the literature. To validate whether the perceptual improvements observed with WD-based losses generalize to other 3DGS frameworks, we additionally integrate WD losses into \emph{Mip-Splatting}~\cite{yu2024mip} and \emph{Scaffold-GS}~\cite{lu2024scaffold}, as well as into the \emph{Comp-GS}~\cite{liu2024compgs} compression scheme.

For fair comparisons, we align the representation budget of WD/WD-R across all compared methods and ensure comparisons are made under similar or lower budgets (\ie fewer splat counts or smaller model size). This is achieved by adjusting the global scale $\gamma$ in~\cref{eq:2D_distortion} while keeping all other hyperparameters fixed. This procedure ensures that differences in performance arise from the optimization objective alone, rather than increased representation capacity.

More details about the training schedules, hyperparameters, and runtime are discussed in~\cref{appendix_sec:training_details}. The WD implementation we adopt~\cite{WD_orig} is unoptimized and incurs a ${\approx}4.5\times$ training overhead over original 3DGS; preliminary tuning (caching ground-truth VGG features and pruning zero-weight VGG pyramid levels) reduces the WD per-iteration time by ${\approx}48\%$ with bit-exact loss and gradients, bringing the relative training overhead down to ${\approx}2.8\times$.

\paragraph{\textbf{Datasets}}
Following the methodology of~\cite{zhou2025perceptual}, we evaluate the effectiveness of our perceptual training across 21 scenes drawn from 4 widely used novel view synthesis datasets: Mip-NeRF 360~\cite{barron2022mipnerf360} (9 scenes), Deep Blending~\cite{DeepBlending2018} (2 scenes), Tanks \& Temples~\cite{Knapitsch2017} (2 scenes), and BungeeNeRF~\cite{xiangli2022bungeenerf} (8 scenes). For Mip-NeRF 360, we distinguish between indoor scenes (\texttt{Counter}, \texttt{Room}, \texttt{Kitchen}, \texttt{Bonsai}) and outdoor scenes (the remaining 5), as these categories involve significantly different numbers of splats and memory footprints. For the same reason, for both our human preference and compression studies, we also report results separately for indoor and outdoor datasets.

\paragraph{\textbf{Perceptual metrics}}
We evaluate the rendering quality using several widely adopted perceptual metrics: LPIPS~\cite{zhang2018unreasonable} and DISTS~\cite{Ding_2020}, Fréchet Inception Distance (FID)~\cite{FID_orig}, and CLIP-based Maximum Mean Discrepancy (CMMD)~\cite{CMMD_orig}. Additional implementation details of perceptual metrics are provided in~\cref{appendix_sec:perceptual_metrics}. For completeness, PSNR and SSIM~\cite{SSIM, MS-SSIM} are also reported in~\cref{appendix_sec:metrics}.

\paragraph{\textbf{Human preference study}}
\label{sec:method_userstudy}
In addition to providing quantitative perceptual metrics, we also conduct a large-scale user study to directly assess perceptual alignment with human preference. The study is carried out on Mabyduck~\cite{Mabyduck}, an independent platform for user preference studies. Following the methodology of the CLIC compression challenge~\cite{clic-challenge}, each trial presents participants with a blind A/B comparison of the same $704\times 704$ random crop of two rendered images from different methods, compared against the ground-truth reference. Participants are asked to choose which rendered image appears closer to the reference. A total of 428 participants completed 39,320 pairwise trials across all the datasets and three frameworks. Responses are aggregated using Bayesian Elo rating based on the Bradley--Terry model~\cite{CaronDoucet2012,Mabyduck-elo}, producing perceptual preference scores for all evaluated methods. More details can be found in~\cref{appendix_sec:human_rater_study}.

\section{Results} \label{sec:results}

To understand the effectiveness of different perceptual objectives under realistic efficiency constraints, we conduct a comprehensive study across multiple 3DGS rendering settings. We first compare the original loss, the composite loss, WD, and WD-R against recent perceptual 3DGS variants---including \emph{Pixel-GS}~\cite{zhang2024pixel} and the recent state-of-the-art splat-efficient perceptual method, \emph{Perceptual-GS}~\cite{zhou2025perceptual}---on standard reconstruction benchmarks under controlled representation budgets. WD-based objectives consistently achieve the best perceptual performance. We then verify that these improvements align with human visual perception through a large-scale human preference study, and provide visualizations for how perceptual optimization affects the learned 3DGS representations. Finally, we demonstrate that WD-based optimization generalizes across methodologically different frameworks---including anti-aliased rendering (\emph{Mip-Splatting}~\cite{yu2024mip}), structured Gaussian representations (\emph{Scaffold-GS}~\cite{lu2024scaffold}), and entropy-constrained compression (\emph{Comp-GS}~\cite{liu2024compgs}).

\subsection{3DGS representation}
\label{subsec:3dgs_representation}
\paragraph{\textbf{Quantitative results}}

\begin{table*}[t]
\centering
  \begin{minipage}[t]{0.49\linewidth}
  \vspace{0pt}
  \centering
  \captionsetup{font=scriptsize, skip=4pt}
  \scriptsize
  \captionof{table}{Perceptual metric results for original 3DGS~\cite{kerbl3Dgaussians} variants on indoor and outdoor datasets. Splat counts are indicated as \# $\mathrm{G}$.}
  \label{tab:3dgs_generation_indoor_outdoor}
  \resizebox{\linewidth}{!}{%
  \begin{tabular}{@{\hspace{1.2pt}}l@{\hspace{2pt}}l@{\hspace{3pt}}c@{\hspace{3pt}}c@{\hspace{3pt}}c@{\hspace{3pt}}c@{\hspace{3pt}}c@{\hspace{1.6pt}}}
  \toprule
  Dataset & Method & \# $\mathrm{G}$ $\downarrow$ & $\mathrm{LPIPS}$ $\downarrow$ & $\mathrm{DISTS}$ $\downarrow$ & $\mathrm{FID}$ $\downarrow$ & $\mathrm{CMMD}$ $\downarrow$ \\
  \midrule
  \multirow{6}{*}{\shortstack{Deep  \\ Blending \\ (Indoors)}}
  & Original loss~\cite{kerbl3Dgaussians} & 2.81M & 0.243 & 0.243 & 106.93 & 0.711 \\
  & Pixel-GS~\cite{zhang2024pixel} & 4.64M & 0.246 & 0.248 & 110.17 & 0.759 \\
  & Perceptual-GS~\cite{zhou2025perceptual} & 2.86M & \seag{0.230} & \seag{0.231} & \seag{93.27} & \seagg{0.586} \\
  \cmidrule{2-7}
  & Composite & 3.96M & 0.235 & 0.240 & 101.11 & 0.765 \\
  & WD & 2.81M & \seagg{0.201} & \seagg{0.205} & \seagg{88.56} & \seaggg{0.584} \\
  & \mbox{WD-R} & 2.87M & \seaggg{0.193} & \seaggg{0.194} & \seaggg{82.32} & \seag{0.606} \\
  \bottomrule
  \addlinespace[2pt]
  \multirow{6}{*}{\shortstack{Mip-NeRF \\ 360  \\ (Indoors)}}
  & Original loss~\cite{kerbl3Dgaussians} & 1.42M & 0.188 & 0.158 & 80.70 & \seag{0.465} \\
  & Pixel-GS~\cite{zhang2024pixel} & 2.49M & 0.177 & 0.147 & 73.36 & \seagg{0.412} \\
  & Perceptual-GS~\cite{zhou2025perceptual} & 1.58M & \seag{0.170} & \seag{0.142} & \seag{69.86} & \seaggg{0.398} \\
  \cmidrule{2-7}
  & Composite & 1.99M & 0.171 & 0.143 & 82.17 & 0.519 \\
  & WD & 1.46M & \seagg{0.152} & \seagg{0.117} & \seaggg{65.59} & 0.511 \\
  & \mbox{WD-R} & 1.49M & \seaggg{0.147} & \seaggg{0.114} & \seagg{67.80} & 0.496 \\
  \bottomrule
  \addlinespace[2pt]
  \multirow{6}{*}{\shortstack{Mip-NeRF \\ 360  \\ (Outdoors)}}
  & Original loss~\cite{kerbl3Dgaussians} & 4.52M & 0.244 & 0.218 & 104.88 & 0.610 \\
  & Pixel-GS~\cite{zhang2024pixel} & 7.40M & \seaggg{0.206} & \seag{0.186} & 65.49 & \seagg{0.416} \\
  & Perceptual-GS~\cite{zhou2025perceptual} & 3.55M & \seaggg{0.206} & 0.188 & \seaggg{58.97} & \seaggg{0.391} \\
  \cmidrule{2-7}
  & Composite & 6.50M & 0.216 & 0.199 & \seagg{59.09} & 0.472 \\
  & WD & 3.54M & 0.228 & \seagg{0.178} & 65.69 & \seag{0.438} \\
  & \mbox{WD-R} & 3.47M & \seaggg{0.206} & \seaggg{0.168} & \seag{59.25} & \seag{0.438} \\
  \bottomrule
  \addlinespace[2pt]
  \multirow{6}{*}{\shortstack{Tanks \\ \& Temples \\ (Outdoors)}}
  & Original loss~\cite{kerbl3Dgaussians} & 1.83M & 0.176 & 0.149 & 53.18 & 1.171 \\
  & Pixel-GS~\cite{zhang2024pixel} & 4.49M & \seag{0.150} & \seag{0.128} & 40.61 & 0.999 \\
  & Perceptual-GS~\cite{zhou2025perceptual} & 1.72M & 0.151 & 0.132 & 40.75 & 0.978 \\
  \cmidrule{2-7}
  & Composite & 1.73M & 0.158 & 0.137 & \seag{39.88} & \seag{0.966} \\
  & WD & 1.70M & \seagg{0.138} & \seagg{0.102} & \seagg{29.69} & \seaggg{0.672} \\
  & \mbox{WD-R} & 1.72M & \seaggg{0.127} & \seaggg{0.096} & \seaggg{29.07} & \seagg{0.737} \\
  \bottomrule
  \addlinespace[2pt]
  \multirow{6}{*}{\shortstack{Bungee\\NeRF \\ (Outdoors)}}
  & Original loss~\cite{kerbl3Dgaussians} & 6.92M & \seag{0.098} & 0.106 & 62.23 & \seag{0.207} \\
  & Pixel-GS~\cite{zhang2024pixel} & \multicolumn{5}{c}{OOM in \texttt{Pompidou} scene} \\
  & Perceptual-GS~\cite{zhou2025perceptual} & 4.97M & \seagg{0.095} & \seag{0.103} & \seag{58.23} & 0.227 \\
  \cmidrule{2-7}
  & Composite & 11.30M & 0.197 & 0.200 & 101.66 & 0.527 \\
  & WD & 4.67M & 0.116 & \seagg{0.100} & \seagg{50.68} & \seagg{0.199} \\
  & \mbox{WD-R} & 4.89M & \seaggg{0.092} & \seaggg{0.087} & \seaggg{46.21} & \seaggg{0.171} \\
  \bottomrule
  \end{tabular}}
  \end{minipage}%
  \hspace{0.01\linewidth}%
  \begin{minipage}[t]{0.49\linewidth}
  \vspace{0pt}
  \centering
  \captionsetup{font=scriptsize, skip=4pt}
  \scriptsize

  \captionof{table}{Perceptual metric comparison for Mip-Splatting~\cite{yu2024mip} and its WD-based variants. WD-based objectives consistently improve the perceptual performance of Mip-Splatting while maintaining similar splat counts (\# $\mathrm{G}$).}
  \label{tab:mipsplatting_mipnerf360}
  \resizebox{\linewidth}{!}{%
  \begin{tabular}{llccccc}
  \toprule
  Dataset & Method & \#G$\downarrow$ & LPIPS$\downarrow$ & DISTS$\downarrow$ & FID$\downarrow$ & CMMD$\downarrow$ \\
  \midrule
  \multirow{3}{*}{\shortstack{Mip-NeRF \\ 360  \\ (Indoors)}}
   & Mip-Splatting & 1.81M & 0.152 & 0.138 & 71.16 & 0.329 \\
   \cmidrule{2-7}
   & +WD          & 1.55M & \seagg{0.134} & \seagg{0.113} & \seagg{54.86} & \seagg{0.301} \\
   & +WD-R        & 1.69M & \seaggg{0.123} & \seaggg{0.106} & \seaggg{53.22} & \seaggg{0.273} \\
  \midrule
  \multirow{3}{*}{\shortstack{Mip-NeRF \\ 360  \\ (Outdoors)}}
   & Mip-Splatting & 5.71M & \seagg{0.193} & 0.180 & 62.54 & 0.327 \\
   \cmidrule{2-7}
   & +WD          & 5.68M & 0.202 & \seagg{0.161} & \seagg{49.78} & \seagg{0.316} \\
   & +WD-R        & 5.37M & \seaggg{0.181} & \seaggg{0.153} & \seaggg{45.87} & \seaggg{0.267} \\
  \bottomrule
  \end{tabular}}

 \vspace{11.5pt}

  \captionof{table}{Perceptual metric comparison for Scaffold-GS~\cite{lu2024scaffold} and its WD-based variants. Model size is reported in MB. WD-based objectives consistently improve perceptual metrics without increasing model sizes.}
  \label{tab:scaffold_mipnerf360}
  \resizebox{\linewidth}{!}{%
  \begin{tabular}{llccccc}
  \toprule
  Dataset & Method & Model Size$\downarrow$ & LPIPS$\downarrow$ & DISTS$\downarrow$ & FID$\downarrow$ & CMMD$\downarrow$ \\
  \midrule
  \multirow{3}{*}{\shortstack{Mip-NeRF \\ 360  \\ (Indoors)}}
   & Scaffold-GS & 99.5\,MB  & 0.166 & 0.150 &  75.08 & \seaggg{0.391} \\
   \cmidrule{2-7}
   & +WD         & 98.4\,MB  & \seagg{0.143} & \seaggg{0.117} & \seaggg{64.41} & \seagg{0.402} \\
   & +WD-R       & 93.9\,MB  & \seaggg{0.140} & \seagg{0.118} & \seagg{67.20} & 0.426 \\
  \midrule
  \multirow{3}{*}{\shortstack{Mip-NeRF \\ 360  \\ (Outdoors)}}
   & Scaffold-GS & 220.7\,MB & 0.273 & 0.255 & 133.33 & 0.686 \\
   \cmidrule{2-7}
   & +WD         & 215.8\,MB & \seagg{0.245} & \seagg{0.202} & \seaggg{67.32} & \seaggg{0.536} \\
   & +WD-R       & 223.5\,MB & \seaggg{0.233} & \seaggg{0.198} & \seagg{67.74} & \seagg{0.559} \\
  \bottomrule
  \end{tabular}}

  \end{minipage}

\end{table*}

\cref{tab:3dgs_generation_indoor_outdoor} provides a comprehensive evaluation of the studied perceptual loss families across indoor and outdoor datasets. We report both perceptual quality metrics and the average splat count per dataset ($\#\mathrm{G}$), allowing methods to be compared under similar representation budgets. We observe that the WD-based losses (WD and \mbox{WD-R}) consistently and substantially outperform the original loss, while also yielding more compact representations (\eg reducing the Gaussian count on the BungeeNeRF dataset from 6.92M to 4.89M). Notably, they also outperform Perceptual-GS~\cite{zhou2025perceptual}, the state-of-the-art in terms of perceptual metrics to date, across the vast majority of perceptual metrics and datasets that we evaluate.

\paragraph{\textbf{Human preference study}}
To verify that improvements in perceptual metrics translate to perceived visual quality, we conduct a large-scale human preference study as described in~\cref{sec:method_userstudy}. In this study, a total of 320 participants completed 30,720 pairwise trials across all the datasets. We evaluate WD, \mbox{WD-R}, Perceptual-GS~\cite{zhou2025perceptual}, Pixel-GS~\cite{zhang2024pixel}, and the original 3DGS loss~\cite{kerbl3Dgaussians} across both indoor and outdoor scenes. The composite loss is excluded from the human preference study, as it incurs significantly higher Gaussian counts on several datasets, making direct comparison less meaningful.

As shown in~\cref{fig:elo_indoor_outdoor}, \mbox{WD-R} achieves significantly better Elo scores across all scenes. The difference in Elo of more than 150 as compared with the original loss implies that the \mbox{WD-R} reconstructions were chosen by raters $2.3\times$ as often. Compared with the state-of-the-art method, Perceptual-GS, both WD and WD-R achieve better Elo scores (within the $95\%$ error margin). Specifically, based on an Elo difference of $72$, \mbox{WD-R} was preferred more than $1.5\times$ over Perceptual-GS.

Overall, the \mbox{WD-R} loss is consistently preferred by participants over all other methods. These results align with the metrics reported in~\cref{tab:3dgs_generation_indoor_outdoor}, confirming that WD-based optimization produces the most perceptually appealing novel views for both indoor and outdoor scenes.

\begin{figure}[t]
  \begin{minipage}[b]{0.66\linewidth}
  \centering
  \includegraphics[width=\linewidth, trim=0 12 0 0]{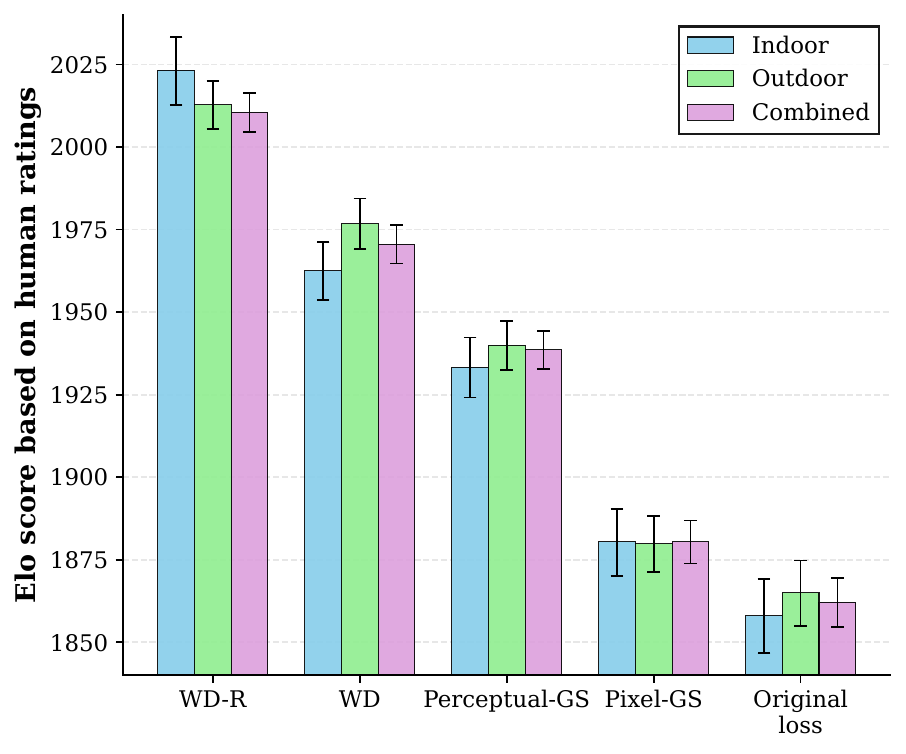}
      \captionof{figure}{Bayesian Elo scores for 3DGS representation methods across indoor, outdoor, and all scenes combined. \mbox{WD-R} and WD achieve the highest scores in all settings (within the 95\% confidence interval).} 
      \label{fig:elo_indoor_outdoor}
  \end{minipage}%
  \hfill%
  \begin{minipage}[b]{0.31\linewidth}
      \centering
      \refstepcounter{figure}
      \begin{subfigure}{\linewidth}
          \centering
          \includegraphics[width=\linewidth, trim=0 0 0 46, clip]{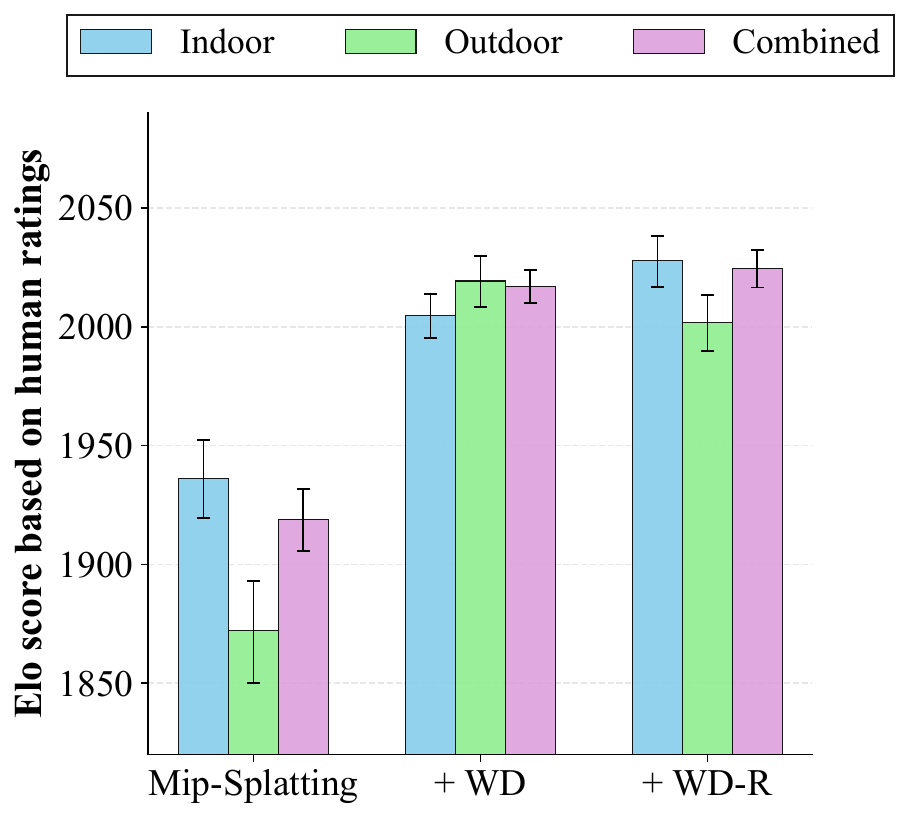}
          \caption{Mip-Splatting}
          \label{fig:mipsplatting_mipnerf360}
      \end{subfigure}

      \vspace{0.4em}

      \begin{subfigure}{\linewidth}
          \centering
          \includegraphics[width=\linewidth, trim=0 0 0 46, clip]{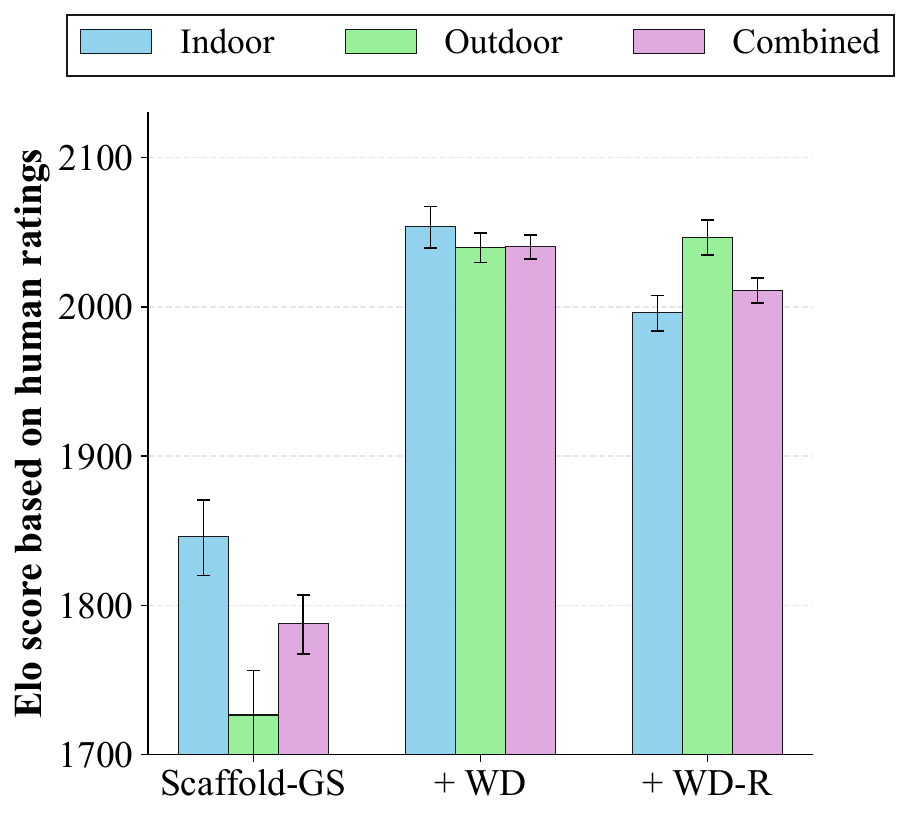}
          \caption{Scaffold-GS}
          \label{fig:scaffold_mipnerf360}
      \end{subfigure}
      \addtocounter{figure}{-1}
      \captionof{figure}{Bayesian Elo scores for (a) Mip-Splatting and (b) Scaffold-GS variants on Mip-NeRF 360~\cite{barron2022mipnerf360}.}
      \label{fig:elo_mip_scaffold}
  \end{minipage}%
\end{figure}

\begin{figure*}[p]
\centering
\begin{tikzpicture}[
    x=.22\linewidth,
    y=-.165\linewidth,
    every node/.style={font=\footnotesize},
]
\node[right, inner sep=0pt] (preview) at (0, 0)
  {\includegraphics[width=.21\linewidth]{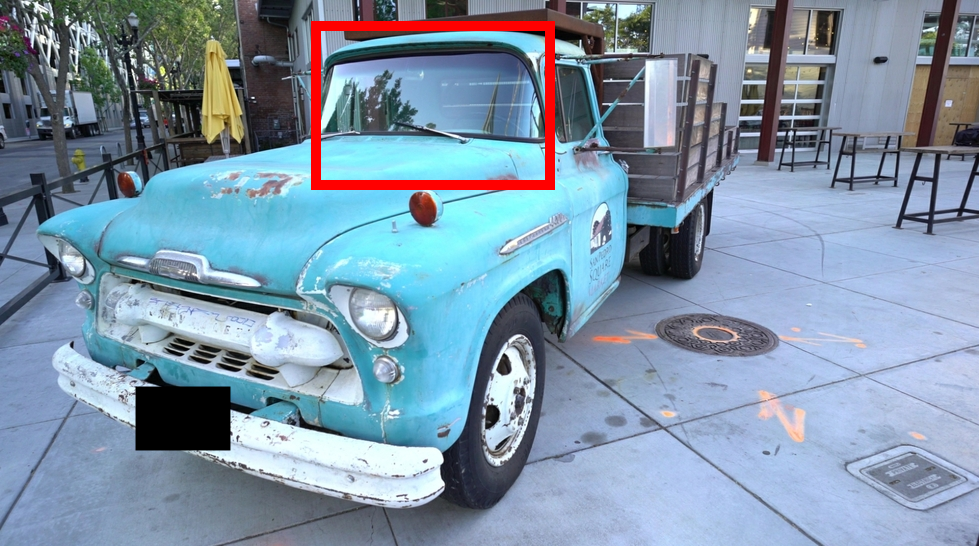}};
\node[left, inner sep=0pt] (gt) at (2, 0)
  {\includegraphics[width=.21\linewidth]{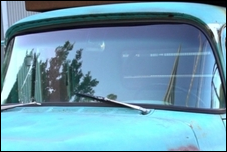}};
\node[left, inner sep=0pt] (orig) at (3, 0)
  {\includegraphics[width=.21\linewidth]{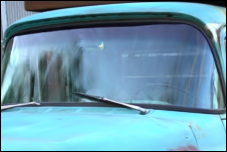}};
\node[left, inner sep=0pt] (wd) at (4, 0)
  {\includegraphics[width=.21\linewidth]{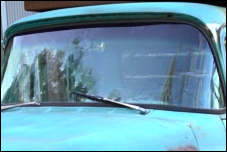}};
\node[left, inner sep=0pt] (pixelgs) at (2, 1.33)
  {\includegraphics[width=.21\linewidth]{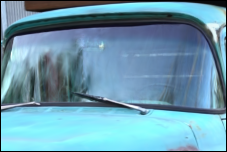}};
\node[left, inner sep=0pt] (perceptualgs) at (3, 1.33)
  {\includegraphics[width=.21\linewidth]{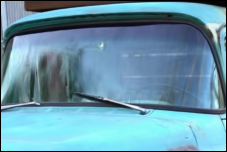}};
\node[left, inner sep=0pt] (wd_orig) at (4, 1.33)
  {\includegraphics[width=.21\linewidth]{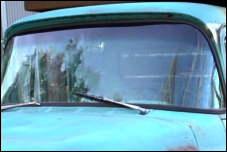}};

\node[below right=0 of preview.south west] {\texttt{Truck}};
\node[below right=0 of gt.south west] {Ground truth};
\node[below right=0 of orig.south west] {\shortstack[l]{Original loss~\cite{kerbl3Dgaussians} \\ (\#G: 2.58M)}};
\node[below right=0 of wd.south west] {\shortstack[l]{WD \\ (\#G: 1.89M)}};
\node[below right=0 of pixelgs.south west] {\shortstack[l]{Pixel-GS~\cite{zhang2024pixel} \\ (\#G: 5.18M)}};
\node[below right=0 of perceptualgs.south west] {\shortstack[l]{Perceptual-GS~\cite{zhou2025perceptual}  \\ (\#G: 2.05M)}};
\node[below right=0 of wd_orig.south west] {\shortstack[l]{WD-R \\ (\#G: 1.99M)}};
\end{tikzpicture}

\vspace{0.6em}

\begin{tikzpicture}[
    x=.22\linewidth,
    y=-.165\linewidth,
    every node/.style={font=\footnotesize},
]
\node[right, inner sep=0pt] (preview) at (0, 0)
  {\includegraphics[width=.21\linewidth]{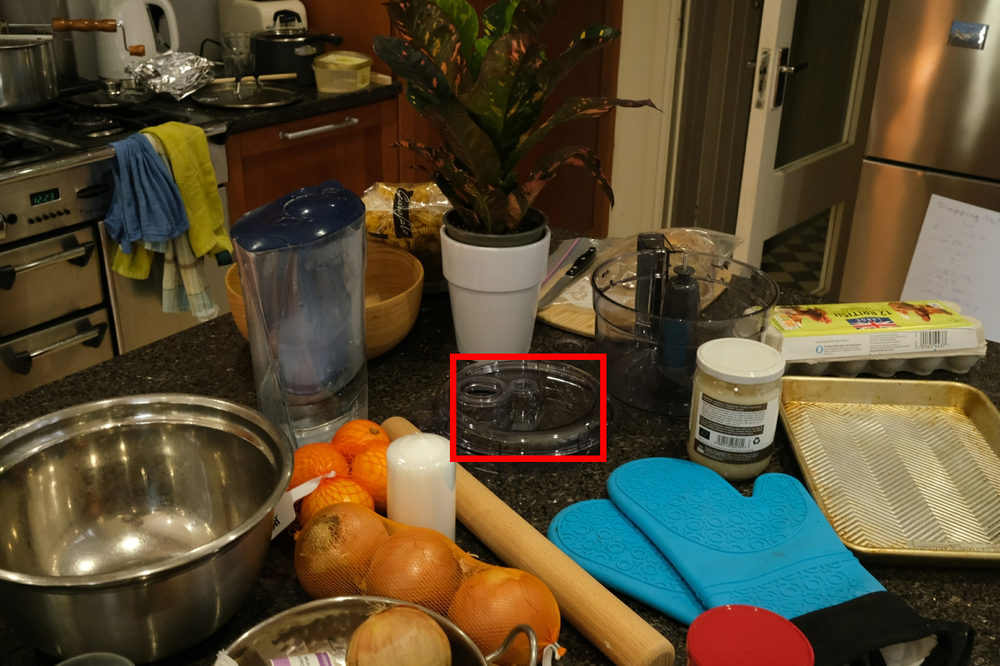}};
\node[left, inner sep=0pt] (gt) at (2, 0)
  {\includegraphics[width=.21\linewidth]{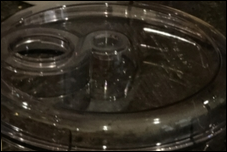}};
\node[left, inner sep=0pt] (orig) at (3, 0)
  {\includegraphics[width=.21\linewidth]{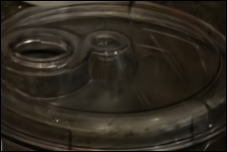}};
\node[left, inner sep=0pt] (wd) at (4, 0)
  {\includegraphics[width=.21\linewidth]{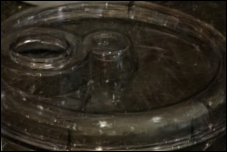}};
\node[left, inner sep=0pt] (pixelgs) at (2, 1.33)
  {\includegraphics[width=.21\linewidth]{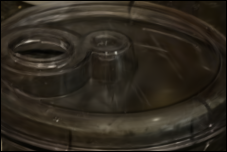}};
\node[left, inner sep=0pt] (perceptualgs) at (3, 1.33)
  {\includegraphics[width=.21\linewidth]{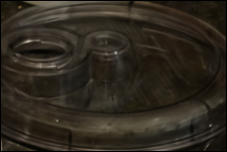}};
\node[left, inner sep=0pt] (wd_orig) at (4, 1.33)
  {\includegraphics[width=.21\linewidth]{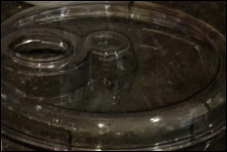}};

\node[below right=0 of preview.south west] {\texttt{Counter}};
\node[below right=0 of gt.south west] {Ground truth};
\node[below right=0 of orig.south west] {\shortstack[l]{Original loss \\  (\#G: 1.17M)}};
\node[below right=0 of wd.south west] {\shortstack[l]{WD  \\ (\#G: 1.10M)}};
\node[below right=0 of pixelgs.south west] {\shortstack[l]{Pixel-GS \\ (\#G: 2.50M)}};
\node[below right=0 of perceptualgs.south west] {\shortstack[l]{Perceptual-GS \\ (\#G: 1.49M)}};
\node[below right=0 of wd_orig.south west] {\shortstack[l]{WD-R \\ (\#G: 1.14M)}};
\end{tikzpicture}

\vspace{0.6em}

\begin{tikzpicture}[
    x=.22\linewidth,
    y=-.165\linewidth,
    every node/.style={font=\small},
]
\node[right, inner sep=0pt] (preview) at (0, 0)
  {\includegraphics[width=.21\linewidth]{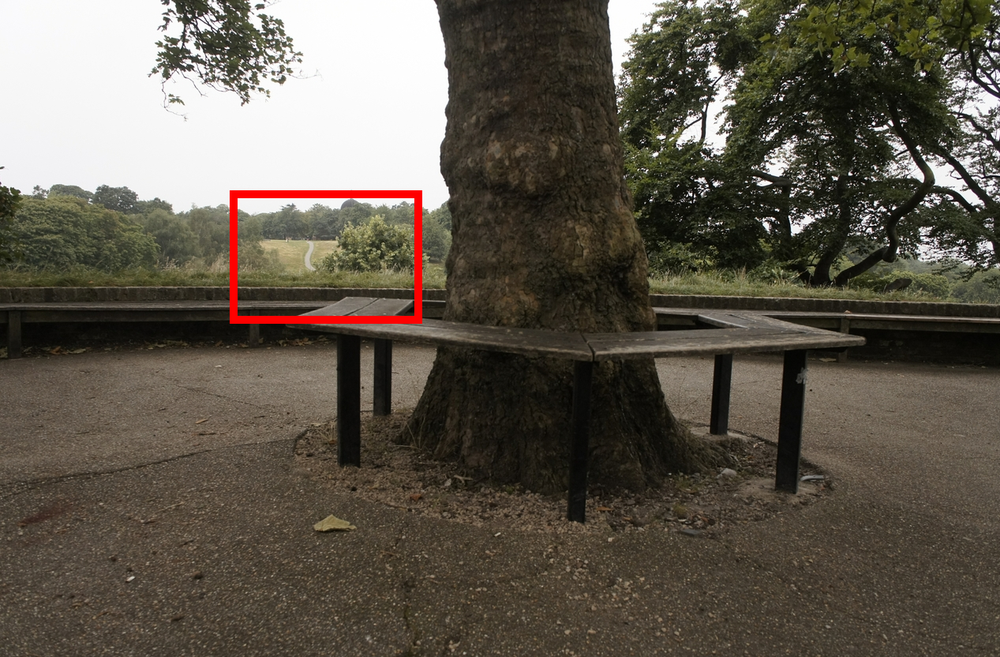}};
\node[left, inner sep=0pt] (gt) at (2, 0)
  {\includegraphics[width=.21\linewidth]{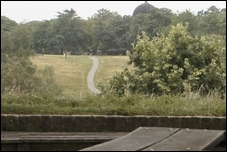}};
\node[left, inner sep=0pt] (orig) at (3, 0)
  {\includegraphics[width=.21\linewidth]{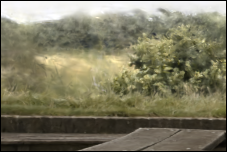}};
\node[left, inner sep=0pt] (wd) at (4, 0)
  {\includegraphics[width=.21\linewidth]{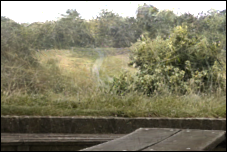}};
\node[left, inner sep=0pt] (pixelgs) at (2, 1.33)
  {\includegraphics[width=.21\linewidth]{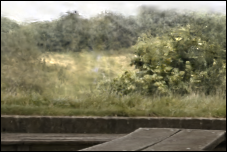}};
\node[left, inner sep=0pt] (perceptualgs) at (3, 1.33)
  {\includegraphics[width=.21\linewidth]{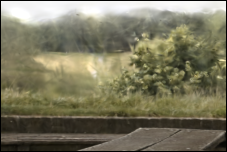}};
\node[left, inner sep=0pt] (wd_orig) at (4, 1.33)
  {\includegraphics[width=.21\linewidth]{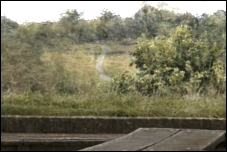}};
\node[below right=0 of preview.south west] {\texttt{Treehill}};
\node[below right=0 of gt.south west] {Ground truth};
\node[below right=0 of orig.south west] {\shortstack[l]{Original loss \\ (\#G: 3.62M)}};
\node[below right=0 of wd.south west] {\shortstack[l]{WD \\ (\#G: 3.72M)}};
\node[below right=0 of pixelgs.south west] {\shortstack[l]{Pixel-GS \\(\#G: 7.47M)}};
\node[below right=0 of perceptualgs.south west] {\shortstack[l]{Perceptual-GS \\ (\#G: 3.48M)}};
\node[below right=0 of wd_orig.south west] {\shortstack[l]{WD-R \\
 (\#G: 3.51M)}};
\end{tikzpicture}

\caption{Visual comparison of the novel view synthesis results obtained by the original 3DGS~\cite{kerbl3Dgaussians}, Pixel-GS~\cite{zhang2024pixel}, Perceptual-GS~\cite{zhou2025perceptual}, and the perceptual loss families discussed in~\cref{subsec:perceptual_losses}. The left images show the full scenes, with detailed crops highlighting reconstruction differences across methods, where $\#\mathrm{G}$ indicates the number of Gaussian splats for each method.}
\label{fig:generation_comparison_main_text}
\end{figure*}

\paragraph{\textbf{Visual results}} \cref{fig:teaser,fig:generation_comparison_main_text} provide a visual comparison of different methods on challenging real-world scenes. The WD loss demonstrates a clear advantage in recovering fine-grained textures and structural details. In the \texttt{Barcelona} scene (\cref{fig:teaser}), WD achieves sharper reconstructions of intricate details on the \emph{Sagrada Família} cathedral, faithfully capturing the small circular patterns on the tower, whereas Perceptual-GS tends to overemphasize linear structures (likely due to its traditional edge detection mechanism derived from Sobel operator). We see similar observations across various datasets as shown in~\cref{fig:generation_comparison_main_text}. Additional visual results are provided in~\cref{appendix_sec:representation_visualization}.

\paragraph{\textbf{Anisotropy under WD}}
As a post-hoc analysis, we study how perceptual optimization affects the geometric properties of the learned Gaussian representations. In particular, we examine the \emph{anisotropy} of the Gaussians. Following~\cite{hyung2024effective}, we measure it using the effective rank (\emph{``erank''}) of the covariance matrix of each Gaussian. For a Gaussian $G_k$ with covariance $\boldsymbol{\Sigma}_k$, its erank is computed from the singular values $s_1^2 \geq s_2^2 \geq s_3^2 > 0$ of $\boldsymbol{\Sigma}_k$ as follows $
\text{erank}(G_k) = \exp\Bigl\{-\!\sum_{i=1}^{3} q_i \log q_i\Bigr\}$, where~$\quad q_i = \frac{s_i^2}{\sum_{j=1}^{3} s_j^2}$.
Lower values ($\approx1$) indicate more anisotropic, needle-like shapes, while higher values ($\approx3$) correspond to isotropic Gaussians.

\cref{fig:erank_hist_barcelona} shows the erank histograms for the \texttt{Barcelona} scene (more details on the other scenes in the BungeeNeRF dataset are in~\cref{appendix_sec:erank_details}). WD-based objectives (WD and \mbox{WD-R}) produce densities shifted towards lower erank values, indicating a stronger tendency to form anisotropic Gaussians compared to pixel-level losses. This allocation pattern reflects how perceptual optimization exploits directional primitives to represent local structures. Specifically, the increased anisotropy allows splats to better represent high-frequency details and fine textures, as their elongated shapes conform to local geometry and reduce rendering blur~\cite{kerbl3Dgaussians, huang20242dgs}. Visual inspection (\cref{fig:generation_comparison_main_text}) and the user study (\cref{fig:elo_indoor_outdoor}) confirm that such anisotropy indeed helps with representing texture and regular structure, for instance in the large-scale city scene example shown in~\cref{fig:teaser}. Beyond Gaussian shape, \mbox{WD-R} also shifts \emph{where} capacity is allocated across the view: \cref{fig:density_heatmap_main} compares splat density heatmaps of original 3DGS and \mbox{WD-R} on a representative view, showing that \mbox{WD-R} concentrates splats on texture-rich regions (a larger version and additional details are provided in~\cref{appendix_sec:density_heatmaps}).

\begin{figure}[t]
    \centering
    \begin{minipage}[t]{0.55\linewidth}
        \vspace{0pt}
        \includegraphics[width=\linewidth]{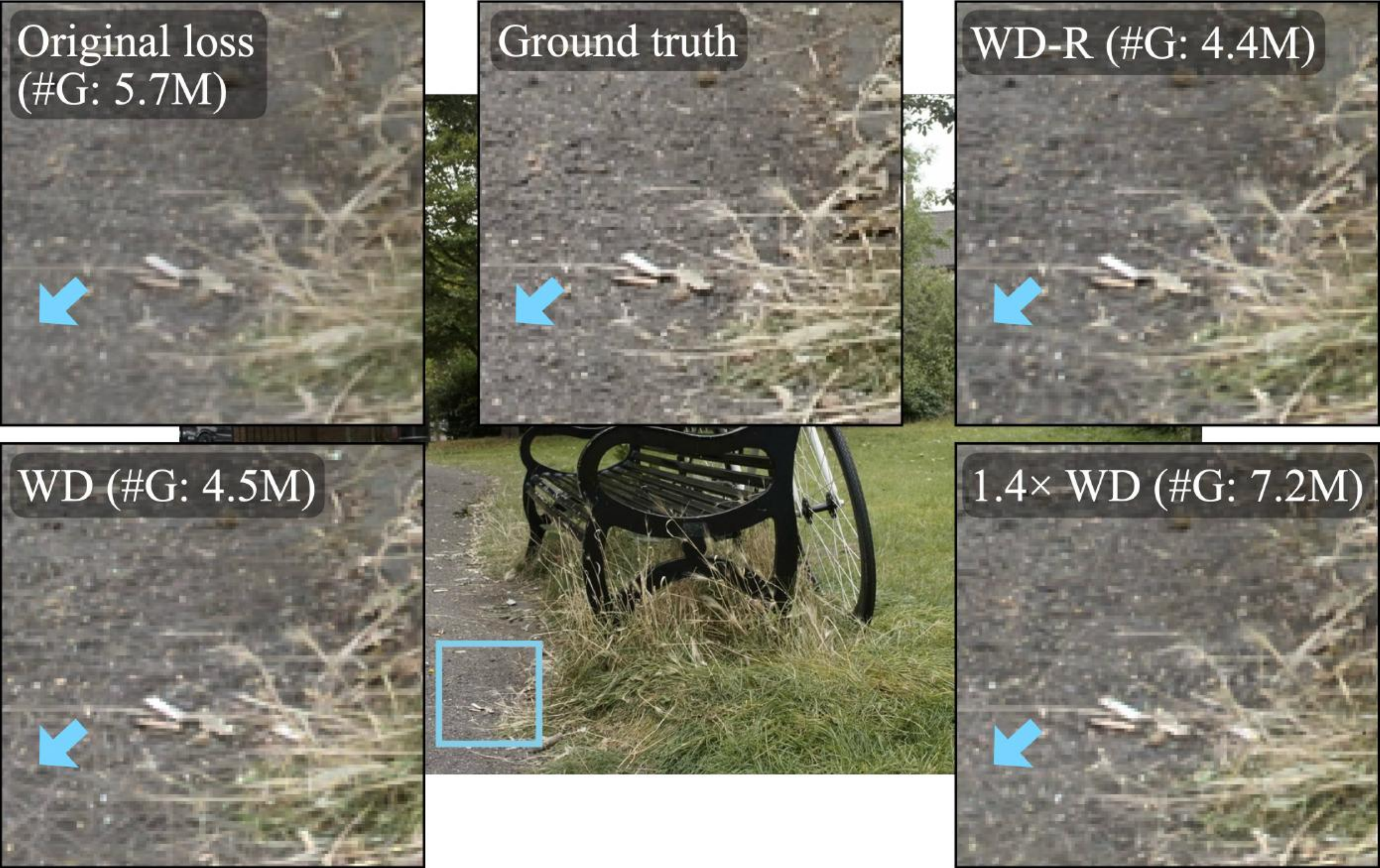}
    \end{minipage}%
    \hfill
    \begin{minipage}[t]{0.43\linewidth}
        \vspace{0pt}
        \caption{Artifacts on \texttt{Bicycle}. With fewer splats (4.5M), WD may produce web-like artifacts (indicated by the \textcolor{arrow}{blue} arrow) imitating texture statistics, while the original loss over-smooths textures. Adding a modest pixel-level term (WD-R) suppresses these without incurring more splats (4.4M). Increasing the global scaling factor ($1.4\times$) of WD also mitigates artifacts with sharper local detail, but raises splat count (7.2M).}
        \label{fig:spiderwebs}
    \end{minipage}
\end{figure}

However, we notice that this tendency can also lead to undesirable ``web-like'' artifacts in some scenarios as shown in~\cref{fig:spiderwebs}, particularly in regions containing dense high-frequency detail but with a low budget of splats. In this situation, WD may allocate a few highly elongated Gaussians to match local statistics (mean and standard deviation in a deep feature space; see~\cref{eq:WD}). This leads to web-like structures that crudely (but economically!) imitate texture. Pixel-level losses avoid this behavior, but instead produce overly smoothed reconstructions.

These artifacts of the WD loss can be mitigated by increasing the splat count, \eg by adjusting the global scaling factor ($\gamma$ in~\cref{eq:2D_distortion}). We find that a more efficient remedy is to use the regularized variant \mbox{WD-R}, which combines WD with a modest pixel-level term. \mbox{WD-R} suppresses these artifacts without inflating the splat count and consistently yields better perceptual metrics (\cref{tab:3dgs_generation_indoor_outdoor}) and human preference ratings (\cref{fig:elo_indoor_outdoor}).

\subsection{Generalization across alternative 3DGS representations}
\label{subsec:3dgs_generalization}

Beyond standard 3DGS, we further investigate whether WD-based optimization remains effective when integrated into architecturally different 3DGS frameworks, specifically Mip-Splatting~\cite{yu2024mip} and Scaffold-GS~\cite{lu2024scaffold}. For both methods, we keep the original framework unchanged and modify only the training objective (\ie~\cref{eq:2D_distortion}), allowing us to isolate the effect of perceptual optimization within each framework.

Mip-Splatting~\cite{yu2024mip} introduces multi-scale filtering to address aliasing artifacts. \cref{tab:mipsplatting_mipnerf360} shows that incorporating WD-based objectives consistently improves perceptual metrics. These improvements are also reflected in the human preference study (\cref{fig:mipsplatting_mipnerf360}), where a total of 4,880 votes were gathered from 86 participants. In particular, \mbox{WD-R} was preferred $1.8\times$ over Mip-Splatting (as suggested by Elo difference of 105.7).

Scaffold-GS~\cite{lu2024scaffold} reduces correlation among Gaussians via structured parameterization, and reports compactness via model size rather than explicit splat counts. \cref{tab:scaffold_mipnerf360} shows similar improvements from WD-based objectives. The human preference study (\cref{fig:scaffold_mipnerf360}), which gathered 3,720 votes from 93 participants, shows that WD-R was preferred $3.6\times$ over Scaffold-GS (Elo difference of 223.2).

As illustrated in~\cref{fig:viz_mipsplatting_scaffoldgs}, WD-R-based losses consistently recover finer textures across both Mip-Splatting and Scaffold-GS variants, largely due to texture resampling, while maintaining comparable resource usage (\#G for Mip-Splatting, model size for Scaffold-GS). Overall, these results suggest that WD-based perceptual optimization is complementary to architectural choices in 3DGS frameworks, and can provide immediate perceptual gains without any modification to the underlying framework. Additional comparisons are provided in~\cref{appendix_sec:3dgs_generalization}.
\begin{figure}[t]
  \vspace*{-0.9em}
  \begin{minipage}[b]{0.40\linewidth}
      \centering
      \begin{subfigure}{0.49\linewidth}\centering
          \includegraphics[width=\linewidth, trim=0 121.6 119.8 7, clip]{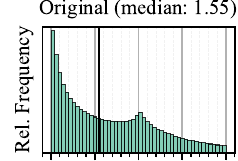}
      \end{subfigure}\hfill
      \begin{subfigure}{0.49\linewidth}\centering
          \includegraphics[width=\linewidth, trim=0 33 119.8 95.8, clip]{figures/erank_hist/barcelona_density_histogram.pdf}
      \end{subfigure}
      \captionof{figure}{\emph{erank}~\cite{hyung2024effective} statistics for \texttt{Barcelona}, comparing original 3DGS~\cite{kerbl3Dgaussians} and \mbox{WD-R}. The black line refers to the median.}
      \label{fig:erank_hist_barcelona}
  \end{minipage}%
  \hspace{0.02\linewidth}%
  \begin{minipage}[b]{0.58\linewidth}
  \centering
    \begin{subfigure}{0.32\linewidth}\centering
        \includegraphics[width=\linewidth]{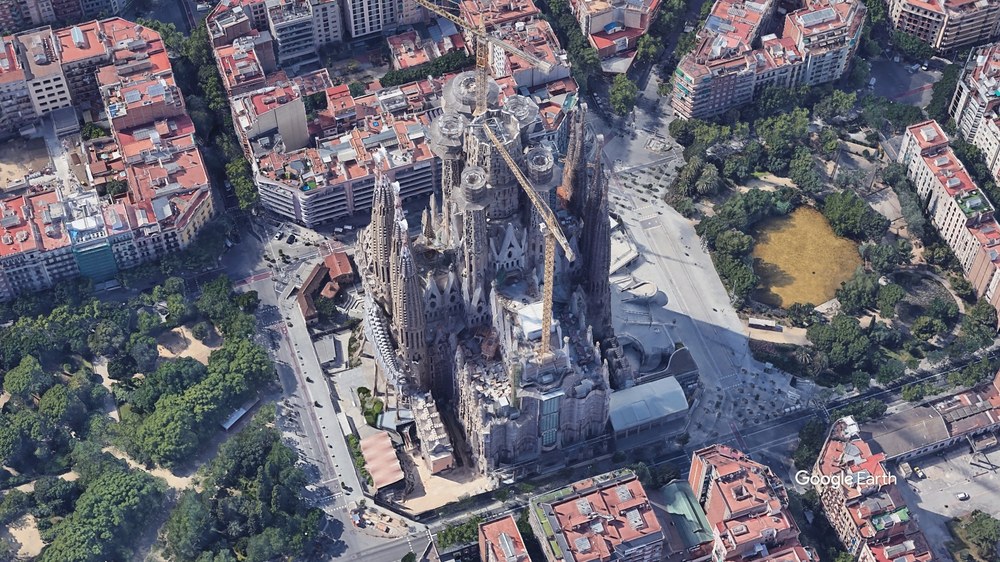}
        \caption{\texttt{Barcelona}.}
    \end{subfigure}
    \begin{subfigure}{0.32\linewidth}\centering
        \includegraphics[width=\linewidth]{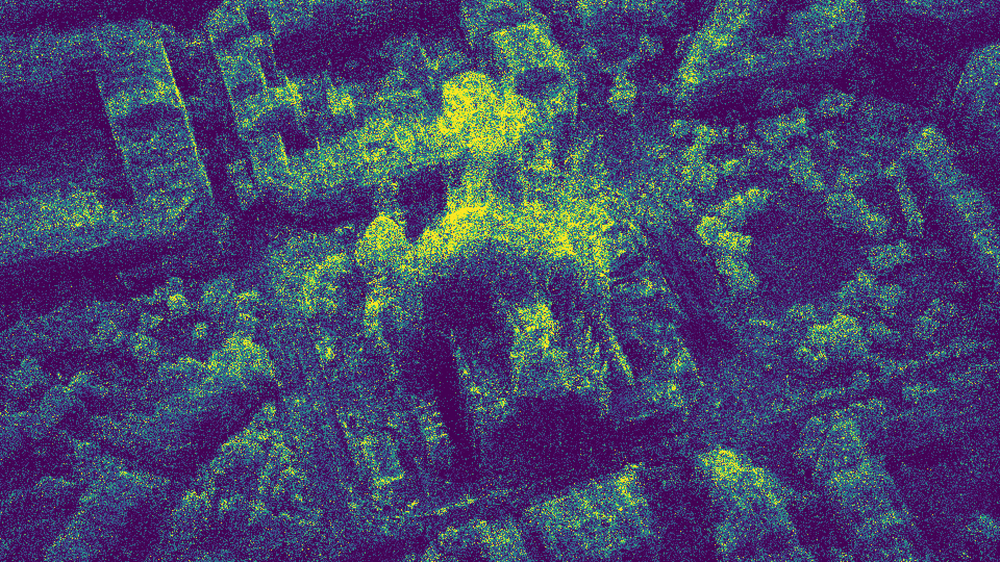}
        \caption{Orig. 3DGS}
    \end{subfigure}
    \begin{subfigure}{0.32\linewidth}\centering
        \includegraphics[width=\linewidth]{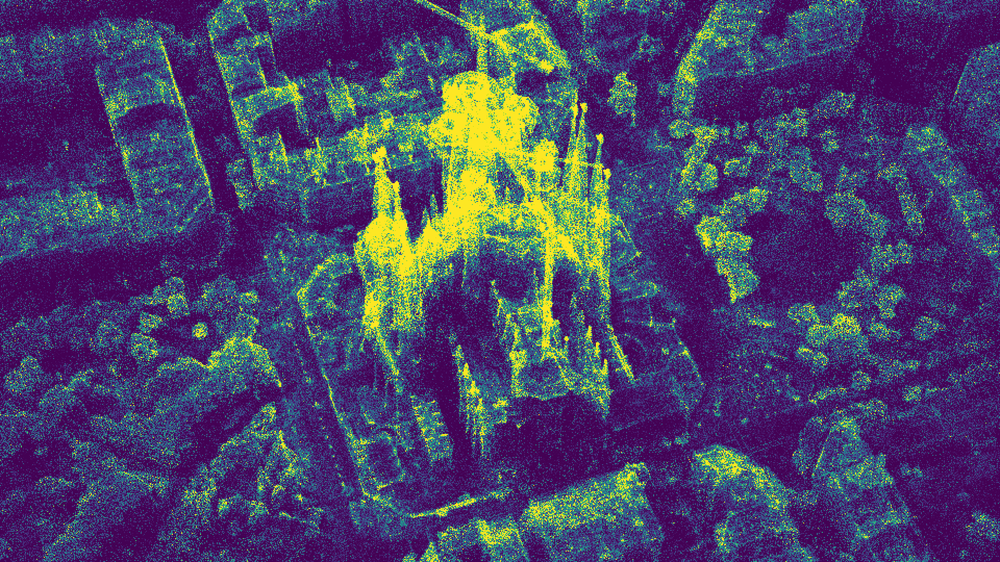}
        \caption{\mbox{WD-R} (ours).}
    \end{subfigure}
    \caption{Splat density heatmaps for \texttt{Barcelona}. \mbox{WD-R} concentrates representation capacity on the texture-rich cathedral, while original 3DGS~\cite{kerbl3Dgaussians} spreads splats more uniformly across the view.}
    \label{fig:density_heatmap_main}
  \end{minipage}
\end{figure}

  \begin{figure*}[t]
\centering
\begin{tikzpicture}[
    x=.2\linewidth,
    y=-.19\linewidth,
    every node/.style={font=\footnotesize},
]
\node[right, inner sep=0pt] (preview) at (0, 0)
  {\includegraphics[width=.18\linewidth]{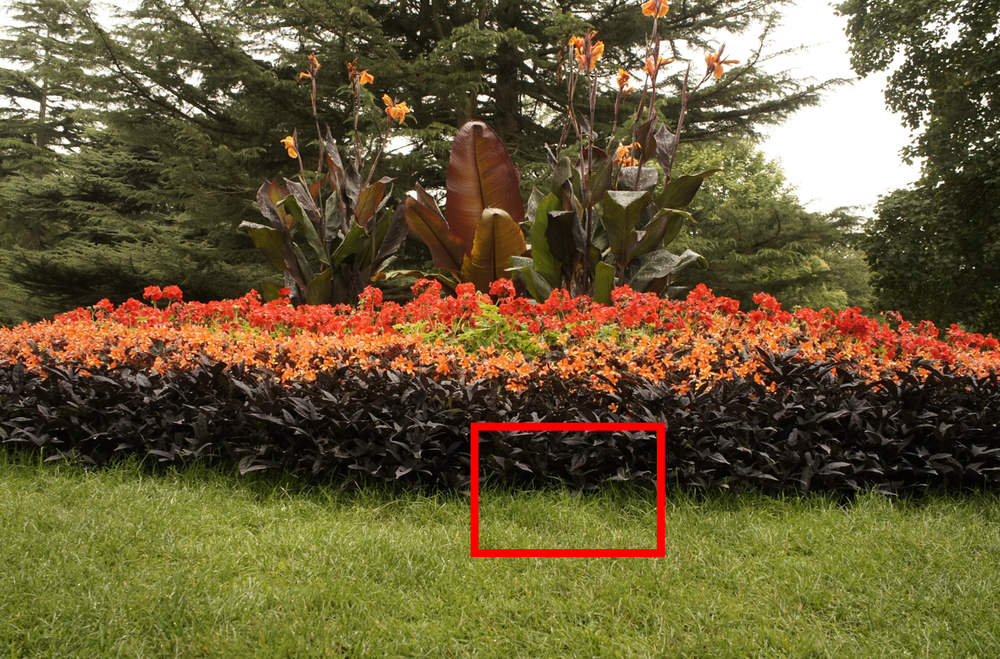}};
\node[left, inner sep=0pt] (gt) at (1.93 , 0)
  {\includegraphics[width=.195\linewidth]{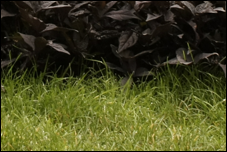}};
\node[left, inner sep=0pt] (orig) at (2.93, 0)
  {\includegraphics[width=.195\linewidth]{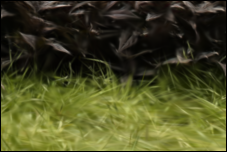}};
\node[left, inner sep=0pt] (wd) at (3.93, 0)
  {\includegraphics[width=.195\linewidth]{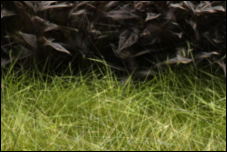}};
  \node[left, inner sep=0pt] (wd_r) at (4.93, 0)
  {\includegraphics[width=.195\linewidth]{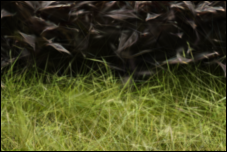}};
\node[below right=0 of preview.south west] {\texttt{Flowers}};
\node[below right=0 of gt.south west] {Ground truth};
\node[below right=0 of orig.south west] {\shortstack[l]{Mip-Splatting\\ \hspace{-.5em}\cite{yu2024mip} (\#G: 4.40M)}};
\node[below right=0 of wd.south west] {\shortstack[l]{+ WD \\ (\#G: 4.22M)}};
\node[below right=0 of wd_r.south west] {\shortstack[l]{+ WD-R \\ (\#G: 4.12M)}};
\end{tikzpicture} 
\begin{tikzpicture}[
    x=.2\linewidth,
    y=-.19\linewidth,
    every node/.style={font=\footnotesize},
]
\node[right, inner sep=0pt] (preview) at (0, 0)
  {\includegraphics[width=.18\linewidth]{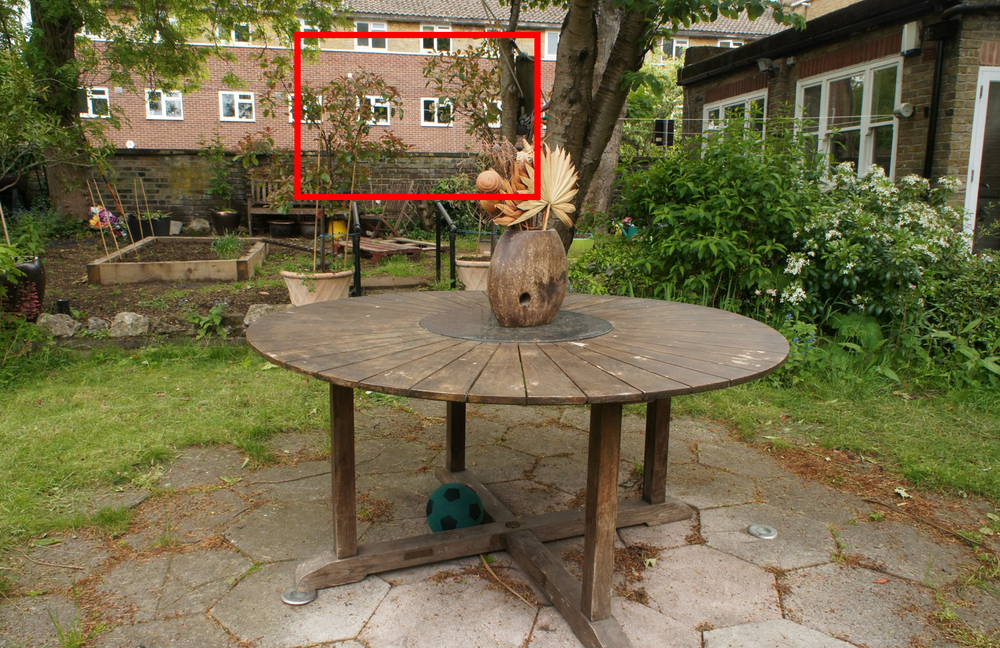}};
\node[left, inner sep=0pt] (gt) at (1.93, 0)
  {\includegraphics[width=.195\linewidth]{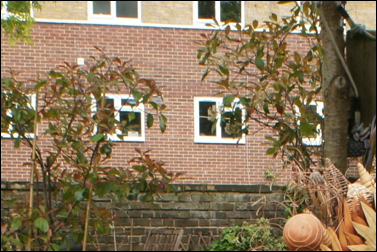}};
\node[left, inner sep=0pt] (orig) at (2.93, 0)
  {\includegraphics[width=.195\linewidth]{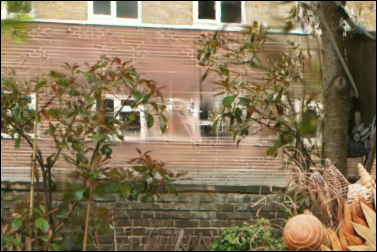}};
\node[left, inner sep=0pt] (wd) at (3.93, 0)
  {\includegraphics[width=.195\linewidth]{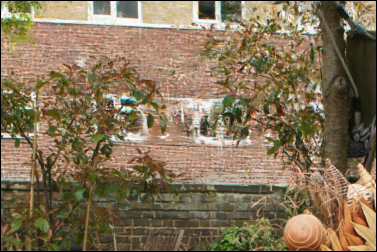}};
  \node[left, inner sep=0pt] (wd_r) at (4.93, 0)
  {\includegraphics[width=.195\linewidth]{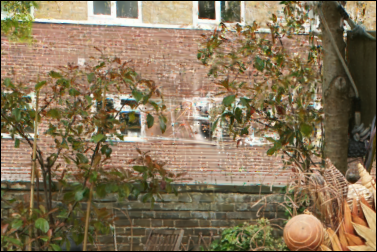}};
\node[below right=0 of preview.south west] {\texttt{Garden}};
\node[below right=0 of gt.south west] {Ground truth};
\node[below right=0 of orig.south west] {\shortstack[l]{Scaffold-GS~\cite{lu2024scaffold} \\ (216.9 MB)}};
\node[below right=0 of wd.south west] {\shortstack[l]{+ WD \\ (168.2 MB)}};
\node[below right=0 of wd_r.south west] {\shortstack[l]{+ WD-R \\ (175.8 MB)}};
\end{tikzpicture}
\caption{Visual comparison of the Mip-Splatting~\cite{yu2024mip}, Scaffold-GS~\cite{lu2024scaffold} and their WD-based variants. The left images show the full scenes, with detailed crops highlighting reconstruction differences across methods, where $\#\mathrm{G}$ indicates the number of Gaussian splats and MB indicates the model size determined by the number of anchors.}
\label{fig:viz_mipsplatting_scaffoldgs}
\end{figure*}

\subsection{Generalization to variable-rate 3DGS compression}
\label{subsec:3dgs_compression}
\begin{figure}[t]
\begin{minipage}{0.8\linewidth}
\centering
    \includegraphics[width=\linewidth]{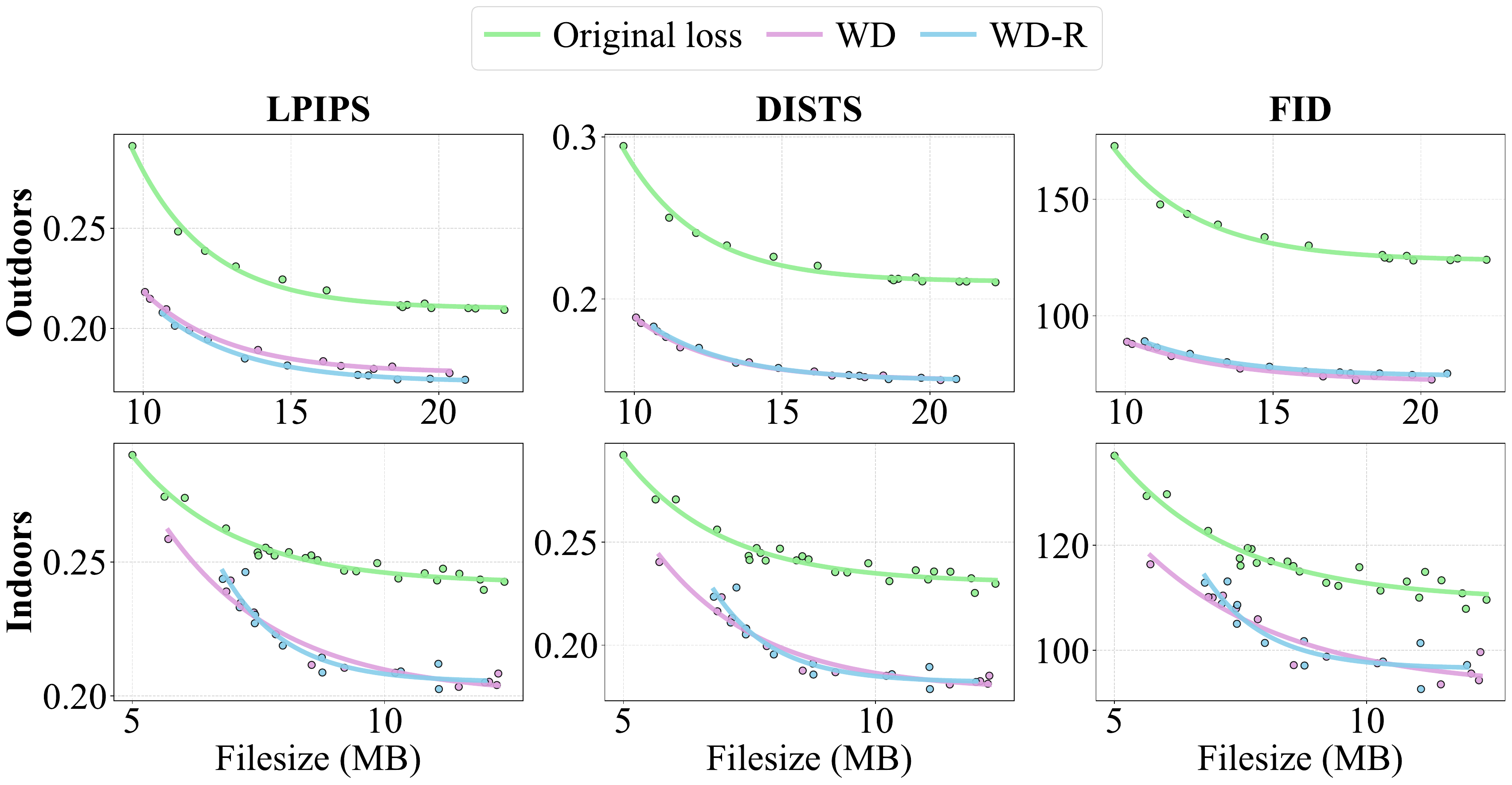}

\end{minipage}
\hfill
\begin{minipage}{0.18\linewidth}
    \caption{Rate--distortion performance of the studied perceptual losses (\cref{subsec:perceptual_losses}) on popular perceptual metrics, using the Comp-GS~\cite{liu2024compgs} compression algorithm.}
    \label{fig:rate_distortion}
\end{minipage}
\end{figure}

Finally, we evaluate WD-based optimization in a compression setting, where training adds explicit entropy coding, yielding a rate--distortion (RD) trade-off as in~\cref{eq:RD}. Here optimization is constrained by the rate term, requiring perceptual gains without added bitrate.~\cref{fig:rate_distortion} shows RD curves for indoor and outdoor scenes, with fitted exponential curves; we sweep $\lambda$ in~\cref{eq:RD} over $[3^{-2}, 3^{-5}]$. Aligned with~\cref{subsec:3dgs_representation,subsec:3dgs_generalization}, WD-based approaches deliver the best perceptual scores (LPIPS, DISTS, FID) across the entire RD region, translating to $\approx 50\%$ bitrate savings at comparable perceptual quality. Visual comparisons are in~\cref{appendix_subsec:compression_visualization}.

\section{Discussion}
Our results demonstrate that a perceptual loss function defined on 2D rendered images can serve as the main objective for reconstructing 3DGS representations from multi-view images. WD, a recently introduced distortion metric focusing on texture perception, and specifically our proposed \mbox{WD-R} variant, achieve state-of-the-art perceptual results in novel view synthesis, as confirmed by LPIPS, DISTS, FID, and CMMD scores, as well as our large-scale human preference study, which to our knowledge is the \mbox{first-of-its-kind} for 3DGS.

It is interesting that WD alone is perfectly suitable as a distortion loss for 2D images within the context of learned image compression and causes no noticeable training instabilities~\cite{WD_orig}, but this appears not to be entirely the case for 3D scene optimization. It can lead to artifacts such as those in~\cref{fig:spiderwebs}, particularly (i)~in regions of 3D space with few training samples, and (ii)~when splat count is highly constrained. Regularizing WD with a modest amount of the original loss (\mbox{WD-R}) suppresses these artifacts and produces substantially better human ratings than all other methods we are aware of. Understanding and addressing the root cause remains an open question.

We also explored spatially varying pooling kernel sizes (\ie $\sigma$) for WD (\cref{appendix_sec:adaptive_sigma}). While these adaptive variants can occasionally improve the reconstruction of structured textures (\eg English text), they overall yield similar perceptual metrics to the $\sigma=4$ case and introduce additional complexity. Future work may further investigate alternative spatially varying $\sigma$ strategies---potentially offering a more targeted remedy for WD artifacts---as well as multi-resolution evaluation with $\sigma$ adapted accordingly.

Additional promising directions include adversarial losses---though their computational complexity would exceed even that of computing classifier features for WD---as well as fully end-to-end optimized approaches to 3DGS where the splat count is explicitly incorporated into the optimization objective.

\clearpage
\newpage

\bibliographystyle{splncs04}
\bibliography{main}

\clearpage
\newpage

\appendix
\section*{Appendix}

\section{Implementation details} \label{appendix_sec:training_details}
We follow the experimental setup of prior work~\cite{kerbl3Dgaussians, zhou2025perceptual}. All experiments are conducted on a single NVIDIA A100 GPU with 80~GB of memory.

\subsection{Runtime comparison of WD and original loss}

We analyze the computational overhead introduced by WD during training. At a rendered resolution of $1063\times1600$ and with the Gaussian count fixed at 4.79M on the \texttt{Bicycle} scene from the Mip-NeRF 360 dataset~\cite{barron2022mipnerf360}, the forward and backward pass time per iteration is 61.1\,ms for the original $\mathrm{L1}+\mathrm{SSIM}$ objective and 273.9\,ms when using WD. In this work, we prioritize perceptual fidelity over raw training efficiency, trading additional training cost for improved visual quality. This $4.5\times$ increase mainly stems from the additional feature extraction and statistical computation in the VGG space (see~\cref{eq:WD}), following the WD implementation in~\cite{WD_orig}. The reference WD implementation is unoptimized, but two simple changes already remove a substantial fraction of this overhead while preserving the loss value and its gradients \emph{bit-exactly} relative to the reference implementation: (i)~caching ground-truth VGG features and their pyramid statistics, since the ground-truth view does not change across training iterations; and (ii)~pruning zero-weight VGG pyramid levels, since the per-level weight $\mathtt{ReLU}(1 - |\log_{2}\sigma - i|)$ is zero for most levels under our constant $\sigma=4$---in our setup only 20 of 96 pyramid levels across the VGG feature stack contribute to the loss. Together, these two changes reduce the WD per-iteration time by ${\approx}48\%$, bringing the relative training overhead down from ${\approx}4.5\times$ to ${\approx}2.8\times$ over original 3DGS. Further VGG-side optimizations (\eg lower-resolution feature extraction, parallelization, or mixed-precision arithmetic) are likely to yield additional speedups and are left to future work. Moreover, WD-based optimization often converges to fewer splats at comparable perceptual quality, which can reduce rendering cost and partially offset the increased training time for the loss computation.


\subsection{Hyperparameters for WD-based loss}
\label{appendix_subsec:loss_hyperparam}
\begin{figure}[h]
      \centering
      \begin{minipage}{0.6\linewidth}                                                                    \includegraphics[width=\linewidth]{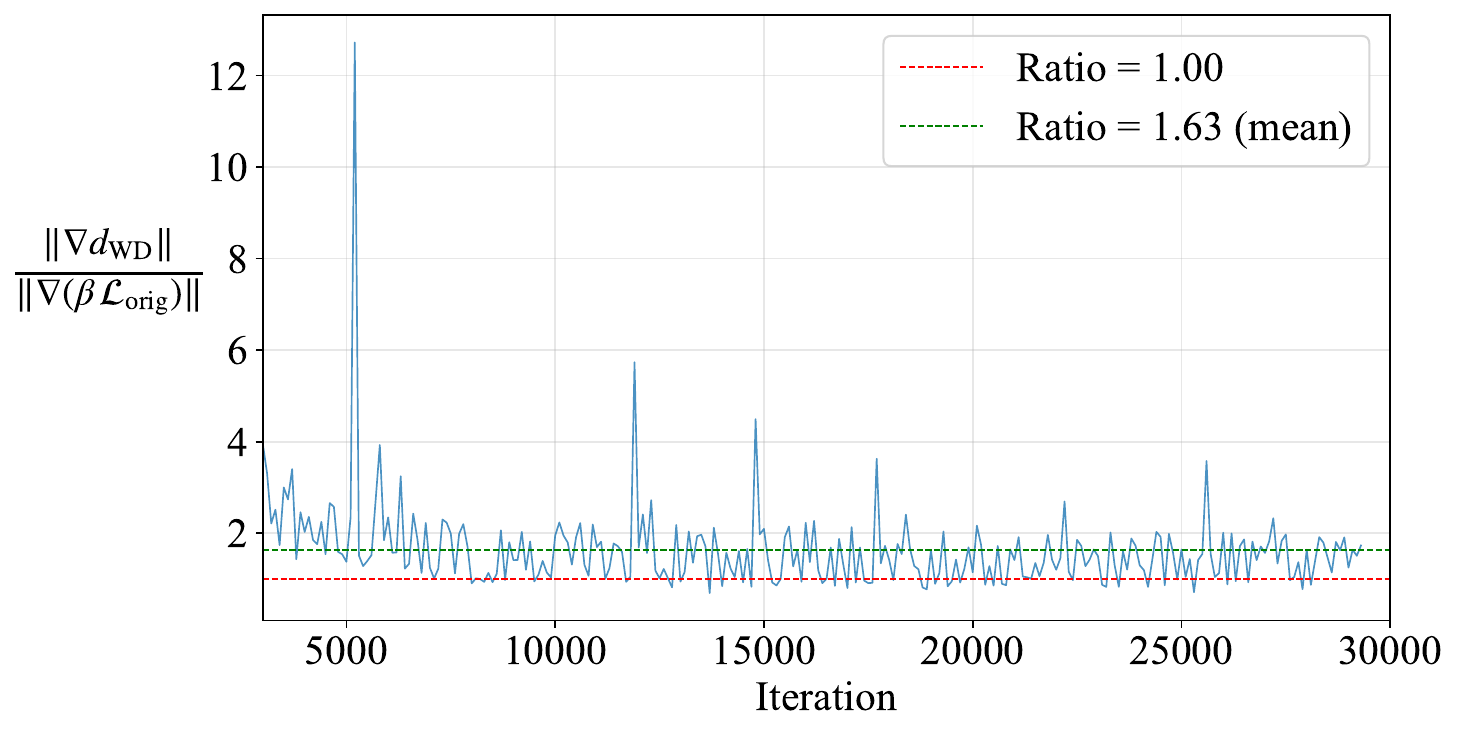}
      \end{minipage}                                                                                     \hfill
      \begin{minipage}{0.38\linewidth}
          \caption{The gradient ratio of $\tfrac{\|\nabla d_{\text{WD}}\|}{\|\nabla (\beta \mathcal{L}_{\text{orig}})\|}$ over training iterations. Despite $\beta$ appearing large, the WD gradient consistently exceeds the scaled original gradient ($\beta \mathcal{L}_{\text{orig}}$) (\ie ratio $> 1$), with a mean ratio of ${\sim}1.6\times$, indicating that $\mathcal{L}_{\text{orig}}$ serves as a moderate regularizer within the WD-R objective.}
          \label{fig:grad_norm}
      \end{minipage}
  \end{figure}

For all 3DGS representation experiments, we follow the default 3DGS training regime~\cite{kerbl3Dgaussians} but apply a dataset-specific global scaling factor $\gamma$ (see~\eqref{eq:2D_distortion}) to each WD-based loss (WD and \mbox{WD-R}, see~\cref{subsec:perceptual_losses}). This is critical because the \textit{adaptive densification} mechanism in the original 3DGS implementation~\cite{kerbl3Dgaussians} relies on the magnitude of positional gradients~\cite{kerbl3Dgaussians}; scaling the global loss magnitude directly scales these gradients, thereby modulating the number of generated Gaussian splats. We tune $\gamma$ to ensure that the resulting Gaussian counts allow for a fair comparison to the baselines. Note that $\gamma$ only affects the overall gradient scale without altering the internal composition of the objective. We use $\beta = \tfrac{1}{0.09}$ for all scenes and choose the global scale $\gamma$ according to the scene so that (i)~splat counts remain comparable to the WD-only baseline, and (ii)~the original loss component $\mathcal{L}_{\text{orig}}$ provides a regularizing influence without overwhelming the WD term. Although $\beta$ may appear large, we empirically verify that the WD term still dominates the gradient signal: on the \texttt{Bicycle} scene for example, the gradient ratio, $\tfrac{\|\nabla d_{\text{WD}}\|}{\|\nabla (\beta \mathcal{L}_{\text{orig}})\|}$, has a mean of ${\sim}1.6\times$ across training iterations (see~\cref{fig:grad_norm}), confirming that $\mathcal{L}_{\text{orig}}$ acts as a moderate regularizer rather than overwhelming the primary WD objective. The exact $\gamma$ values are shown in~\cref{tab:wd_weight}.

\begin{table}[bp]
\centering
\caption{
Global scale $\gamma$ for WD-based losses, \ie WD-only and WD-R. The objectives are defined in \cref{subsec:perceptual_losses}.
Coefficients $\gamma$ are calibrated per dataset to maintain splat counts comparable to the respective baselines, with the original loss $\mathcal{L}_{\text{orig}}$ acting as a mild regularizer.
}
\label{tab:wd_weight}
\begin{tabular}{cc|cc|cc|cc|cc}
\toprule
 \multicolumn{2}{c|}{\shortstack{Deep Blending \\ (Indoors)}}
& \multicolumn{2}{c|}{\shortstack{Mip-NeRF 360 \\ (Indoors)}}
& \multicolumn{2}{c|}{\shortstack{Mip-NeRF 360 \\ (Outdoors)}}
& \multicolumn{2}{c|}{\shortstack{Tanks \& Temples \\ (Outdoors)}}
& \multicolumn{2}{c}{\shortstack{BungeeNeRF \\ (Outdoors)}} \\
\midrule
WD & WD-R
& WD & WD-R
& WD & WD-R
& WD & WD-R
& WD & WD-R \\
\midrule
0.028 & 0.025
& 0.029 & 0.025
& 0.035 & 0.028
& 0.038 & 0.032
& 0.030 & 0.025 \\
\bottomrule
\end{tabular}
\end{table}

\subsection{Warm-up strategy}
For the \mbox{WD} loss, we warm up the training with the original loss~\cite{kerbl3Dgaussians} for the first 3k iterations before introducing the perceptual objective. This setting is used for all datasets except BungeeNeRF~\cite{xiangli2022bungeenerf}, where we use a 5k-iteration warm-up. This is because BungeeNeRF is a large-scale cityscape dataset derived from Google Earth imagery, featuring large variations in viewing distance and viewpoint, as well as inconsistent appearance (\eg lighting) and geometry, making the optimization relatively less stable. This warm-up stage provides a stable geometric initialization and mitigates instability caused by pruning and densification heuristics, which we hypothesize is due to the WD loss lacking a direct pixel-level fidelity term. Note that such warm-up strategies are commonly used when training with perceptual or adversarial objectives in the literature (\eg~\cite{isola2017image,ledig2017photo}), where an initial pixel-level objective helps stabilize early training. A deeper investigation (\eg eliminating warm-up or using smoother schedules) is a worthwhile direction for future work.
\subsection{Perceptual metrics}
\label{appendix_sec:perceptual_metrics}
LPIPS~\cite{zhang2018unreasonable} and DISTS~\cite{Ding_2020} are computed per image and averaged across validation views. In contrast, FID~\cite{FID_orig} and CMMD~\cite{CMMD_orig} measure differences between feature distributions of rendered and ground-truth images. Specifically, FID uses Inception-v3 features, while CMMD is computed using CLIP-ViT-L/14-336 embeddings. As FID and CMMD require sufficiently large sample sets for stable estimation, we augment the evaluation data by extracting five spatial crops per view and also adding their horizontal flips. The metrics are then computed per scene and averaged across all scenes to ensure statistical stability while accounting for inter-scene variation.

It is worth noting that FID and CMMD operate at the distribution level, quantifying the divergence between the reconstructed and ground-truth data. As such, they serve as indicators of \textit{realism} rather than perceptual fidelity to a specific reference image. However, it is common practice in perceptual image processing and compression research to report these distributional metrics alongside per-image metrics like DISTS and LPIPS (see~\cite{mentzer2020highfidelitygenerativeimagecompression, muckley_implicit}). Since each dataset we evaluate on only contains a small handful of individual scenes, and hence does not represent a larger ensemble of scenes well, we chose to compare distributions \emph{per scene} (\ie the FID or CMMD value is computed by comparing the distribution of original views from each scene to the distribution of novel rendered views from the model fitted to the same scene).
\section{Ablation on composite loss hyperparameters}
\label{appendix_sec:ablation_composite}

The goal of these ablations is to empirically find a set of weights ($\omega_1$--$\omega_4$) on the L1, L2, MS-SSIM~\cite{MS-SSIM} and LPIPS~\cite{zhang2018unreasonable} losses in~\eqref{eq:composite_loss}, respectively, that achieve a good balance of reconstruction quality as assessed by LPIPS, DISTS, FID, and CMMD, which are commonly used to evaluate image processing algorithms~\cite{mentzer2020highfidelitygenerativeimagecompression, muckley_implicit}.

First, we replace SSIM~\cite{SSIM} with MS-SSIM, its multi-scale counterpart. Second, we introduce L2 and LPIPS, gradually increasing the LPIPS weight $\omega_4$. While doing this, we track our measures of success across two datasets: the outdoor Tanks \& Temples and the indoor Deep Blending (\cref{tab:composite_ablation}). We observe that $\omega_4 = 0.1$ represents a sweet spot: increasing the weight further yields diminishing metric improvements while the splat count remains low.

We further test permutations of the resulting weights, with the result that none of the permutations improve upon the previously identified sweet spot. We stop our search here, and choose $\omega_1=0.05, \omega_2 = 0.30, \omega_3=0.60, \omega_4=0.10$ as the final weighting for the composite loss.

\begin{table}[t]
\scriptsize
\centering
\caption{Ablation study on the hyperparameters of the composite loss in~\cref{eq:composite_loss}.}
\label{tab:composite_ablation}
\begin{tabular}{l|llll|c|cccccc}
\toprule
Dataset & $\omega_{1}$ & $\omega_{2}$ & $\omega_{3}$  & $\omega_{4}$ & \#$\mathrm{G} \downarrow $ & $\mathrm{LPIPS}$  $\downarrow$ & $\mathrm{DISTS}$  $\downarrow$& $\mathrm{FID}$ $\downarrow$ & $\mathrm{CMMD}$ $\downarrow$ & $\mathrm{PSNR}$ $\uparrow$  & $\mathrm{SSIM}$ $\uparrow$ \\
\midrule
\multirow{9}{*}{\shortstack{Deep\\Blending \\ (Indoors)}} &
0.2 & -- & 0.8 & -- & 2.81M & 0.243 & 0.243 & 106.93 & \seagg{0.711} & \seaggg{29.60} & \seaggg{0.903} \\
& 0.05 & 0.3 & 0.6 & -- & 2.80M & \seagg{0.235} & \seagg{0.237} & \seaggg{90.20} & \seaggg{0.602} & \seag{29.47} & \seagg{0.901} \\
& 0.05 & 0.3 & 0.6 & 0.05 & 3.31M & \seaggg{0.233} & \seaggg{0.236} & \seagg{100.90} & \seag{0.744} & 29.30 & \seag{0.896} \\
& 0.05 & 0.3 & 0.6 & 0.10 & 3.96M & \seagg{0.235} & 0.240 & \seag{101.11} & 0.765 & 28.87 & 0.836\\
& 0.05 & 0.3 & 0.6 & 0.20 & 5.75M & 0.242 & 0.253 & 106.66 & 0.808 & 27.62 & 0.875  \\
& 0.05 & 0.3 & 0.6 & 0.40 & 11.05M & 0.353 & 0.379 & 181.54 & 1.393 & 22.40 & 0.764 \\
& 0.05 & 0.3 & 0.6 & 0.80 & 14.95M & 0.566 & 0.589 & 335.53 & 2.728 & 14.13 & 0.529 \\
& 0.05 & 0.6 & 0.3 & 0.1 & 2.50M & \seagg{0.235} & \seag{0.238} & 106.88 & 0.785 & \seagg{29.55} & 0.893 \\
& 0.05 & 0.1 & 0.3 & 0.6 & 14.35M & 0.670 & 0.678 & 408.34 & 3.502 & 11.11 & 0.404 \\
\midrule
\multirow{9}{*}{\shortstack{Tanks \\ \& Temples \\ (Outdoors)}} &
0.2 & -- & 0.8 & -- & 1.83M & 0.176 & 0.149 & 53.18 & 1.171 & \seag{23.63} & \seagg{0.846}\\
& 0.05 & 0.3 & 0.6 & -- & 1.29M & 0.171 & 0.150 & 47.60 & 1.084 & \seaggg{23.69} & \seaggg{0.847} \\
& 0.05 & 0.3 & 0.6 &  0.05 & 1.48M & \seag{0.162} & \seag{0.139}  & 42.13 & 0.997 & 23.51 & \seag{0.842} \\
& 0.05 & 0.3 & 0.6 &  0.1 & 1.73M & \seagg{0.158} & \seagg{0.137} & \seag{39.88} & \seag{0.966} &  23.54 & 0.840 \\
& 0.05 & 0.3 & 0.6 &  0.2 & 2.37M & \seaggg{0.154} & \seaggg{0.135} & \seaggg{36.38} & \seaggg{0.924} & 23.36 & 0.836 \\
& 0.05 & 0.3 & 0.6 &  0.4 & 4.09M & 0.164 & 0.149  & \seagg{37.66} & \seagg{0.963} & 22.80 & 0.813 \\
& 0.05 & 0.3 & 0.6 &  0.8 & 7.39M & 0.272 & 0.267 & 93.44 & 1.699 & 18.60 & 0.670 \\
& 0.05 & 0.6 & 0.3 & 0.1 & 1.12M & 0.172 & 0.149 & 44.12 & 1.054 & \seagg{23.64} & 0.829 \\
& 0.05 & 0.1 & 0.3 & 0.6 & 9.55M & 0.419 & 0.423 & 176.41 & 2.814 & 13.49 & 0.481 \\
\bottomrule
\end{tabular}
\end{table}

\section{Ablation on $\sigma$ choices for Wasserstein Distortion}
\subsection{Constant $\sigma$ map}
\label{appendix_sec:ablation_wd_sigma}
As discussed in~\cref{subsec:perceptual_losses}, $\sigma$ controls the permissiveness to \emph{texture resampling}, where lower values allow less resampling~\cite{WD_Qiu, WD_orig}. The measurements in \cref{tab:wd_sigma_generation} show that there is a trade-off: decreasing $\sigma$, at a constant loss weight $\gamma$, improves pointwise perceptual metrics, as $\sigma=0$ corresponds to pointwise comparisons between feature maps (or pixels). At the same time, this comes at the cost of an increased splat count. For our main results, we selected $\sigma=4$, which would appear to be a suboptimal choice according to the metrics listed here. However, the fact that we largely achieve better human ratings with WD/WD-R than optimizing for combinations of such pointwise metrics, as evidenced by the human preference study in~\cref{fig:elo_indoor_outdoor}, suggests that pointwise metrics simply are not accurately aligned with human texture perception. Although the original WD implementation paper~\cite{WD_orig} highlights a saliency-guided $\sigma$-map, we note that a fixed $\sigma$ for the WD loss remains a strong baseline itself, as demonstrated in~\cite[Sec.~3, Fig.~5]{WD_orig}.

\begin{table}[t]
\scriptsize
\centering
\caption{Perceptual metric comparison across $\sigma$-values for WD on Mip-NeRF 360~\cite{barron2022mipnerf360} dataset.}
\label{tab:wd_sigma_generation}
\begin{tabular}{lcccccccc}
\toprule
{\centering Dataset} & $\sigma$ & \# $\mathrm{G} \downarrow$ & $\mathrm{LPIPS}$ $\downarrow$ & $\mathrm{DISTS}$$\downarrow$ & $\mathrm{FID}$ $\downarrow$ & $\mathrm{CMMD}$ $\downarrow$ & $\mathrm{PSNR}$ $\uparrow$  & $\mathrm{SSIM}$ $\uparrow$ \\
\midrule
\multirow{4}{*}{\parbox{2cm}{\shortstack{Mip-NeRF 360  \\ (Indoors)}}}
& 1 & 4.24M & \seaggg{0.134} & \seaggg{0.104} & \seaggg{56.60} & \seaggg{0.373} & \seaggg{29.82} & \seaggg{0.902} \\
& 2 & 2.72M & \seagg{0.136} & \seagg{0.106} & \seagg{58.88} & \seagg{0.435} & \seagg{29.74} & \seagg{0.891} \\
& 4 & 1.46M & \seag{0.152} & \seag{0.117} & \seag{65.59} & \seag{0.511} & \seag{29.24} & \seag{0.852} \\
& 8 & 0.89M & 0.184 & 0.143 & 83.40 & 0.625 & 28.14 & 0.822 \\
\bottomrule
\addlinespace[2pt]
\multirow{4}{*}{\parbox{2cm}{\shortstack{Mip-NeRF 360  \\ (Outdoors)}}}
& 1 & 12.40M & \seaggg{0.191} & \seagg{0.164} & \seagg{58.81} & \seagg{0.352} & \seaggg{24.45} & \seaggg{0.720} \\
& 2 & 7.56M & \seagg{0.192} & \seaggg{0.158} & \seaggg{51.58} & \seaggg{0.351} & \seagg{24.11} & \seagg{0.684} \\
& 4 & 3.54M & \seag{0.228} & \seag{0.178} & \seag{65.69} & \seag{0.438} & \seag{22.90} & \seag{0.577} \\
& 8 & 1.86M & 0.279 & 0.216 & 85.75 & 0.629 & 21.44 & 0.471 \\
\bottomrule
\end{tabular}
\end{table}

\begin{table}[t]
\centering
\caption{Perceptual comparison for saliency-based adaptive $\sigma$ vs.\ constant $\sigma{=}4$ on the Mip-NeRF 360 dataset~\cite{barron2022mipnerf360}.}
\label{tab:ablation_adaptive_sigma}
\scriptsize
\begin{tabular}{clccccccc}
  \toprule
  Dataset & Method & \# $\mathrm{G}$ $\downarrow$ & $\mathrm{LPIPS}$ $\downarrow$ & $\mathrm{DISTS}$ $\downarrow$ & $\mathrm{FID}$ $\downarrow$ & $\mathrm{CMMD}$ $\downarrow$ &
  $\mathrm{PSNR}$ $\uparrow$ & $\mathrm{SSIM}$ $\uparrow$ \\
  \midrule
  \multirow{2}{*}{\shortstack{Indoors}}
  & $\sigma = 4$ & 1.46M & \seaggg{0.152} & \seaggg{0.117} & \seaggg{65.59} & \seaggg{0.511} & \seaggg{29.24} & \seaggg{0.852} \\
  & adpt.\ $\sigma$ & 1.66M & \seagg{0.155} & \seagg{0.118} & \seagg{70.78} & \seagg{0.522} & \seagg{29.34} & \seagg{0.857} \\
  \midrule
  \multirow{2}{*}{\shortstack{Outdoors}}
  & $\sigma = 4$ & 3.54M & \seagg{0.228} & \seaggg{0.178} & \seaggg{65.69} & \seaggg{0.438} & \seaggg{22.90} & \seaggg{0.577} \\
  & adpt.\ $\sigma$ & 3.65M & \seaggg{0.222} & \seagg{0.173} & \seagg{67.81} & \seagg{0.443} & \seagg{23.16} & \seagg{0.603} \\
  \bottomrule
  \end{tabular}
  \end{table}

\subsection{Adaptive $\sigma$ map}
\label{appendix_sec:adaptive_sigma}
In addition to the fixed pooling kernel used in the main experiments, we also explored spatially adaptive variants where the WD pooling size (\ie $\sigma$) varies across the image. The goal is to better match perceptual sensitivity by allowing smaller pooling regions in visually important areas and larger regions elsewhere to permit more texture resampling while maintaining similar splat count.

\begin{figure*}[tbp]
\centering
\begin{tikzpicture}[
    x=.25\linewidth,
    y=-.19\linewidth,
    every node/.style={font=\footnotesize},
]
\node[right, inner sep=0pt] (preview) at (0, 0)
  {\includegraphics[width=.24\linewidth]{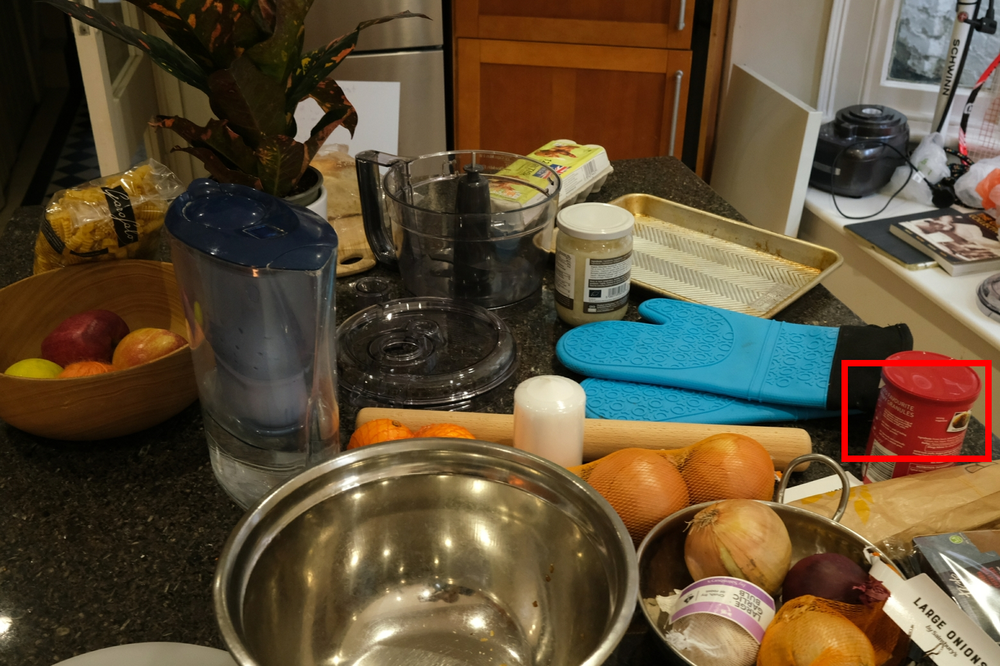}};
\node[left, inner sep=0pt] (gt) at (2, 0)
  {\includegraphics[width=.24\linewidth]{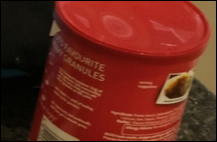}};
\node[left, inner sep=0pt] (orig) at (3, 0)
  {\includegraphics[width=.24\linewidth]{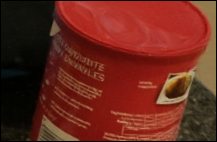}};
\node[left, inner sep=0pt] (wd) at (4, 0)
  {\includegraphics[width=.24\linewidth]{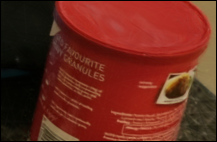}};
\node[below right=0 of preview.south west] {\texttt{Counter}};
\node[below right=0 of gt.south west] {Ground truth};
\node[below right=0 of orig.south west] {\shortstack[l]{+ WD ($\sigma=4$)\\ (\#G: 1.10M)}};
\node[below right=0 of wd.south west] {\shortstack[l]{+ adpt. $\sigma$ \\ (\#G: 1.38M)}};
\end{tikzpicture}
\vspace{-.7cm}
\caption{Visual comparison of constant $\sigma$ vs.\ saliency-guided adaptive $\sigma$, where EML-NET~\cite{JIA20EML} is used to extract saliency maps.}
\label{fig:viz_dynamic_sigma}
\end{figure*}

We investigated two strategies. First, we constructed saliency-guided $\sigma$ maps using EML-NET~\cite{JIA20EML}, assigning smaller $\sigma$ values to highly salient regions and larger values elsewhere. While this approach occasionally improved the reconstruction of fine, structured textures such as English text (see~\cref{fig:viz_dynamic_sigma}), it produced similar aggregate perceptual metrics to the fixed $\sigma$ setting, as shown in~\cref{tab:ablation_adaptive_sigma}.

Second, we explored depth-guided $\sigma$ maps, where the pooling size varies as a function of estimated depth to account for perspective-dependent perceptual sensitivity. We used the monocular depth estimator Depth-Pro~\cite{Bochkovskii2024:arxiv} to obtain per-view depth estimates. In practice, depth-based $\sigma$ maps were more difficult to stabilize during training and did not consistently improve perceptual scores.

Due to the additional complexity and limited quantitative gains, we retain the fixed pooling configuration in the main experiments.

\section{Human rater study}
\label{appendix_sec:human_rater_study}
\begin{figure}[htbp]
\centering
    \includegraphics[width=\linewidth]{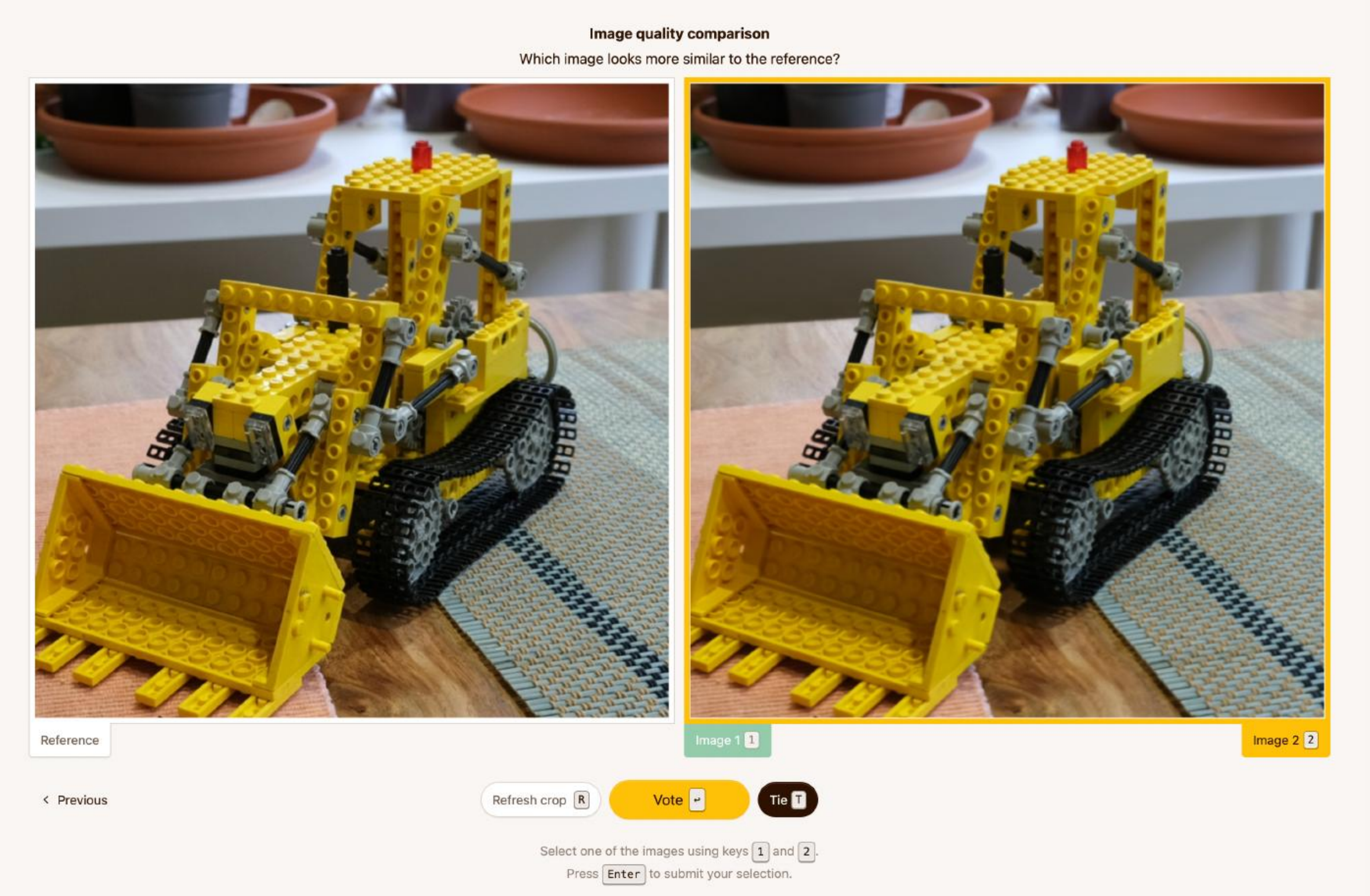}
    \caption{Screenshot of the subjective study interface as seen by the human raters on Mabyduck~\cite{Mabyduck}.}
    \label{fig:user_interface}
\end{figure}

The subjective study is conducted in a blind pairwise comparison format. \cref{fig:user_interface} shows the interface seen by the human raters. The interface allows for blind A/B comparison of two reconstructions (on the right) with the reference image displayed on the left. As in the CLIC compression challenge~\cite{clic-challenge}, the Elo scoring model dynamically chooses which pair of reconstructions (each corresponding to a method) are compared against each other using the maximum information gain strategy~\cite{Mabyduck-strategies} to maximize comparisons which provide a useful signal. Finally, Bayesian Elo scores are computed based on all the pairwise comparisons.

To minimize noisy voting, Mabyduck performs thorough sanity checking of the raters' setup with a pre-screening in accordance with the Ishihara color test~\cite{ishihara1917}. The pre-screening checks for color blindness, contrast sensitivity, and basic ability to detect compression artifacts. A sample pre-screening study is shown here: \url{https://xp.mabyduck.com/en/latest/pre_screen_image/job/j6ne0x2/}.

\section{Additional quantitative reconstruction quality results}

\subsection{3DGS representation}
\label{appendix_sec:metrics}
\cref{tab:3dgs_generation_indoor_outdoor_complete} reports additional metrics (PSNR and SSIM) for the 3DGS representation task discussed in~\cref{subsec:3dgs_representation}. While WD-based losses outperform the original 3DGS objective ($\mathrm{L1}+\mathrm{SSIM}$, \ie original loss)~\cite{kerbl3Dgaussians}, Pixel-GS~\cite{zhang2024pixel}, and Perceptual-GS~\cite{zhou2025perceptual} on popular perceptual metrics (LPIPS, DISTS, FID) and human preference, they yield lower PSNR and SSIM scores. This is likely because baseline methods (original loss in~\cite{kerbl3Dgaussians}, Pixel-GS~\cite{zhang2024pixel}, and Perceptual-GS~\cite{zhou2025perceptual}) explicitly include an SSIM term in their training objectives.

\begin{table}[htbp]
\centering
\scriptsize
\caption{Perceptual metric results for original 3DGS~\cite{kerbl3Dgaussians} variants on indoor and outdoor datasets. Splat counts are indicated as \# $\mathrm{G}$.}
\label{tab:3dgs_generation_indoor_outdoor_complete}
\resizebox{\linewidth}{!}{%
\begin{tabular}{clccccccc}
\toprule
Dataset & Method & \# $\mathrm{G}$ $\downarrow$ & $\mathrm{LPIPS}$ $\downarrow$ & $\mathrm{DISTS}$ $\downarrow$ & $\mathrm{FID}$ $\downarrow$ & $\mathrm{CMMD}$ $\downarrow$ & $\mathrm{PSNR}$ $\uparrow$ & $\mathrm{SSIM}$ $\uparrow$ \\
\midrule
\multirow{6}{*}{\shortstack{Deep  \\ Blending}}
& Original loss~\cite{kerbl3Dgaussians} & 2.81M & 0.243 & 0.243 & 106.93 & 0.711 & \seagg{29.60} & \seagg{0.903} \\
& Pixel-GS~\cite{zhang2024pixel} & 4.64M & 0.246 & 0.248 & 110.17 & 0.759 & 28.98 & \seag{0.896} \\
& Perceptual-GS~\cite{zhou2025perceptual} & 2.86M & \seag{0.230} & \seag{0.231} & \seag{93.27} & \seagg{0.586} & \seaggg{29.94} & \seaggg{0.905} \\
\cmidrule{2-9}
& Composite & 3.96M & 0.235 & 0.240 & 101.11 & 0.765 & 28.95 & 0.837\\
& WD & 2.81M &   \seagg{0.201} & \seagg{0.205} & \seagg{88.56} & \seaggg{0.584} & 28.98 & 0.823 \\
& \mbox{WD-R} & 2.87M & \seaggg{0.193} & \seaggg{0.194} & \seaggg{82.32} & \seag{0.606} & \seag{29.58} & 0.839 \\
\bottomrule
\addlinespace[2pt]
\multirow{6}{*}{\shortstack{Mip-NeRF \\ 360  \\ (Indoors)}}
& Original loss~\cite{kerbl3Dgaussians} & 1.42M & 0.188 & 0.158 & 80.70 & \seag{0.465} & \seag{30.95} & \seag{0.922} \\
& Pixel-GS~\cite{zhang2024pixel} & 2.49M & 0.177 & 0.147 & 73.36 & \seagg{0.412} & \seagg{31.17} & \seagg{0.926} \\
& Perceptual-GS~\cite{zhou2025perceptual} & 1.58M & \seag{0.170} & \seag{0.142} & \seag{69.86} & \seaggg{0.398} & \seaggg{31.39} & \seaggg{0.929} \\
\cmidrule{2-9}
& Composite & 1.99M & 0.171 & 0.143 &  82.17 & 0.519 & 30.52 & 0.889 \\
& WD & 1.46M & \seagg{0.152} & \seagg{0.117} & \seaggg{65.59} & 0.511 & 29.24 & 0.852 \\
& \mbox{WD-R} & 1.49M & \seaggg{0.147} & \seaggg{0.114} & \seagg{67.80} & 0.496 & 30.44 & 0.881 \\
\bottomrule
\addlinespace[2pt]
\multirow{6}{*}{\shortstack{Mip-NeRF \\ 360  \\ (Outdoors)}}
& Original loss~\cite{kerbl3Dgaussians} & 4.52M & 0.244 & 0.218 & 104.88 & 0.610 & 24.27 & \seag{0.708}\\
& Pixel-GS~\cite{zhang2024pixel} & 7.40M & \seaggg{0.206} & \seag{0.186} & 65.49 & \seagg{0.416} & \seagg{24.31} & \seagg{0.720} \\
& Perceptual-GS~\cite{zhou2025perceptual} & 3.55M & \seaggg{0.206} & 0.188 & \seaggg{58.97} & \seaggg{0.391} & \seaggg{24.38} & \seaggg{0.726}\\
\cmidrule{2-9}
& Composite & 6.50M & 0.216 & 0.199 & \seagg{59.09} & 0.472 & \seag{24.28} & 0.699 \\
& WD & 3.54M & 0.228 & \seagg{0.178} & 65.69 & \seag{0.438} & 22.90 & 0.577 \\
& \mbox{WD-R} & 3.47M & \seaggg{0.206} & \seaggg{0.168} & \seag{59.25} & \seag{0.438} & 24.09 & 0.677 \\
\bottomrule
\addlinespace[2pt]
\multirow{6}{*}{\shortstack{Tanks \\ \& Temples}}
& Original loss~\cite{kerbl3Dgaussians} & 1.83M & 0.176 & 0.149 & 53.17 & 1.171 & \seag{23.63} & \seag{0.846}\\
& Pixel-GS~\cite{zhang2024pixel} & 4.49M & \seag{0.150} & \seag{0.128} & 40.61 & 0.999 & \seagg{23.73} & \seagg{0.855}\\
& Perceptual-GS~\cite{zhou2025perceptual} & 1.72M & 0.151 & 0.132 & 40.75 & 0.978 & \seaggg{23.90} & \seaggg{0.856}\\
\cmidrule{2-9}
& Composite & 1.73M & 0.158 & 0.137 & \seag{39.88} & \seag{0.966} &  23.54 & 0.840 \\
& WD & 1.70M & \seagg{0.138} & \seagg{0.102} & \seagg{29.69} & \seaggg{0.672} & 22.53 & 0.742 \\
& \mbox{WD-R} & 1.72M & \seaggg{0.127} & \seaggg{0.096} & \seaggg{29.07} & \seagg{0.737} & 23.47 & 0.791 \\
\bottomrule
\addlinespace[2pt]
\multirow{6}{*}{\shortstack{Bungee\\NeRF}}
& Original loss~\cite{kerbl3Dgaussians} & 6.92M & \seag{0.098} & 0.106 & 62.23 & \seag{0.207} & \seagg{27.67} & \seagg{0.914}\\
& Pixel-GS~\cite{zhang2024pixel} & \multicolumn{7}{c}{OOM in \texttt{Pompidou} scene} \\
& Perceptual-GS~\cite{zhou2025perceptual} & 4.97M & \seagg{0.095} & \seag{0.103} & \seag{58.23} & 0.227 & \seaggg{27.86}& \seaggg{0.918} \\
\cmidrule{2-9}
& Composite & 11.30M & 0.197 & 0.200 & 101.66 & 0.527 & 24.12 & 0.785 \\
& WD & 4.67M & 0.116 & \seagg{0.100} & \seagg{50.68}  & \seagg{0.199} & 25.52 & 0.807 \\
& \mbox{WD-R} & 4.89M & \seaggg{0.092} & \seaggg{0.087} & \seaggg{46.21} & \seaggg{0.171} & \seag{27.45} & \seag{0.881} \\
\bottomrule
\end{tabular}
}
\end{table}

\subsection{Generalization across alternative 3DGS representations}
\label{appendix_sec:3dgs_generalization}

\cref{tab:mipsplatting_more,tab:scaffold_more} report results when integrating WD and WD-R into the other two 3DGS frameworks studied, \emph{Mip-Splatting}~\cite{yu2024mip} and \emph{Scaffold-GS}~\cite{lu2024scaffold}, across additional datasets beyond those in~\cref{subsec:3dgs_generalization}. For most datasets, WD-based objectives consistently improve perceptual metrics such as LPIPS, DISTS, FID, and CMMD relative to the original methods. In particular, WD-R provides the most stable improvements, achieving the best or near-best perceptual scores across the majority of datasets while maintaining comparable resource usage (Gaussian count for Mip-Splatting and model size for Scaffold-GS). As expected for perceptual optimization, these gains are sometimes accompanied by small decreases in PSNR and SSIM, reflecting the known trade-off between pixel-level fidelity and perceptual quality~\cite{blau2018perception}. Overall, the results confirm that the benefits of WD-based training generalize across multiple datasets and extend beyond the original 3DGS algorithm~\cite{kerbl3Dgaussians} to alternative rendering pipelines.

\begin{table}[t]
  \centering
  \caption{Perceptual metric comparison for Mip-Splatting~\cite{yu2024mip} and its WD-based variants across datasets.}
  \label{tab:mipsplatting_more}
  \resizebox{\linewidth}{!}{%
  \begin{tabular}{llcccccccc}
  \toprule
  Dataset & Method & \#G $\downarrow$ & LPIPS $\downarrow$ & DISTS $\downarrow$ & FID $\downarrow$ & CMMD $\downarrow$ & PSNR $\uparrow$ & SSIM $\uparrow$ \\
  \midrule
  \multirow{3}{*}{Deep Blending}
   & Mip-Splatting & 3.5M & 0.239          & 0.242          & 102.01          & 0.660          & \seagg{29.35}  & \seagg{0.902}  \\
   & +WD           & 1.9M & \seagg{0.204}  & \seagg{0.208}  & \seagg{89.38}   & \seaggg{0.567} & 29.32          & 0.874          \\
   & +WD-R         & 2.4M & \seaggg{0.194} & \seaggg{0.201} & \seaggg{82.96}  & \seagg{0.614}  & \seaggg{29.46} & \seaggg{0.890} \\
  \midrule
  \multirow{3}{*}{\shortstack{Mip-NeRF 360  \\ (Indoors)}}
   & Mip-Splatting & 1.8M & 0.152          & 0.138          & 71.16           & 0.329          & \seaggg{31.34} & \seaggg{0.935} \\
   & +WD           & 1.5M & \seagg{0.134}  & \seagg{0.113}  & \seagg{54.86}   & \seagg{0.301}  & 29.10          & 0.886          \\
   & +WD-R         & 1.7M & \seaggg{0.123} & \seaggg{0.106} & \seaggg{53.22}  & \seaggg{0.273} & \seagg{30.30}  & \seagg{0.915}  \\
  \midrule
  \multirow{3}{*}{\shortstack{Mip-NeRF 360  \\ (Outdoors)}}
   & Mip-Splatting & 5.7M & \seagg{0.193}  & 0.180          & 62.54           & 0.327          & \seaggg{25.14} & \seaggg{0.760} \\
   & +WD           & 5.7M & 0.202          & \seagg{0.161}  & \seagg{49.78}   & \seagg{0.316}  & 23.72          & 0.638          \\
   & +WD-R         & 5.4M & \seaggg{0.181} & \seaggg{0.153} & \seaggg{45.87}  & \seaggg{0.267} & \seagg{24.79}  & \seagg{0.729}  \\
  \midrule
  \multirow{3}{*}{Tanks \& Temples}
   & Mip-Splatting & 2.4M & 0.156          & 0.136          & 45.55           & 1.045          & \seaggg{23.82} & \seaggg{0.860} \\
   & +WD           & 1.2M & \seagg{0.148}  & \seagg{0.111}  & \seagg{34.84}   & \seagg{0.827}  & 22.40          & 0.772          \\
   & +WD-R         & 1.5M & \seaggg{0.134} & \seaggg{0.103} & \seaggg{32.81}  & \seaggg{0.825} & \seagg{23.32}  & \seagg{0.825}  \\
  \midrule
  \multirow{3}{*}{BungeeNeRF}
   & Mip-Splatting & 8.8M & \seaggg{0.113} & \seagg{0.122}  & \seagg{74.89}   & \seagg{0.336}  & \seaggg{27.37} & \seaggg{0.909} \\
   & +WD           & 6.3M & 0.156          & 0.131          & 75.64           & 0.456          & 24.04          & 0.768          \\
   & +WD-R         & 7.5M & \seagg{0.121}  & \seaggg{0.111} & \seaggg{64.64}  & \seaggg{0.328} & \seagg{26.08}  & \seagg{0.868}  \\
  \bottomrule
  \end{tabular}%
  }
\end{table}

\begin{table}[t]
\centering
\caption{Perceptual metric comparison for Scaffold-GS~\cite{lu2024scaffold} and its WD-based variants across datasets.}
\label{tab:scaffold_more}
\resizebox{\linewidth}{!}{%
  \begin{tabular}{llccccccc}
  \toprule
  Dataset & Method & Size (MB) $\downarrow$ & LPIPS $\downarrow$ & DISTS $\downarrow$ & FID $\downarrow$ & CMMD $\downarrow$ & PSNR $\uparrow$ & SSIM $\uparrow$ \\
  \midrule
  \multirow{3}{*}{Deep Blending}
   & Scaffold-GS & 54.0  & 0.253          & 0.256          & \seaggg{101.53} & \seaggg{0.612} & \seaggg{30.29} & \seaggg{0.909} \\
   & +WD         & 51.1  & \seagg{0.244}  & \seagg{0.248}  & 119.34          & 0.983          & 28.92          & 0.866          \\
   & +WD-R       & 47.2  & \seaggg{0.228} & \seaggg{0.231} & \seagg{109.02}  & \seagg{0.915}  & \seagg{29.75}  & \seagg{0.887}  \\
  \midrule
  \multirow{3}{*}{\shortstack{Mip-NeRF 360  \\ (Indoors)}}
   & Scaffold-GS & 99.5  & 0.166          & 0.150          & 75.08           & \seaggg{0.391} & \seaggg{31.58} & \seaggg{0.933} \\
   & +WD         & 98.4  & \seagg{0.143}  & \seaggg{0.117} & \seaggg{64.41}  & \seagg{0.402}  & 30.16          & 0.899          \\
   & +WD-R       & 93.9  & \seaggg{0.140} & \seagg{0.118}  & \seagg{67.20}   & 0.426          & \seagg{30.90}  & \seagg{0.917}  \\
  \midrule
  \multirow{3}{*}{\shortstack{Mip-NeRF 360  \\ (Outdoors)}}
   & Scaffold-GS & 220.7 & 0.273          & 0.255          & 133.33          & 0.686          & \seaggg{24.70} & \seaggg{0.717} \\
   & +WD         & 215.8 & \seagg{0.244}  & \seagg{0.202}  & \seaggg{67.32}  & \seaggg{0.536} & 23.26          & 0.615          \\
   & +WD-R       & 223.5 & \seaggg{0.233} & \seaggg{0.198} & \seagg{67.74}   & \seagg{0.559}  & \seagg{24.09}  & \seagg{0.677}  \\
  \midrule
  \multirow{3}{*}{Tanks \& Temples}
   & Scaffold-GS & 77.7  & 0.173          & 0.155          & 48.01           & 1.125          & \seaggg{24.05} & \seaggg{0.854} \\
   & +WD         & 63.8  & \seagg{0.150}  & \seagg{0.117}  & \seagg{30.98}   & \seaggg{0.868} & 22.76          & 0.782          \\
   & +WD-R       & 67.8  & \seaggg{0.136} & \seaggg{0.109} & \seaggg{29.60}  & \seagg{0.890}  & \seagg{23.56}  & \seagg{0.831}  \\
  \midrule
  \multirow{3}{*}{BungeeNeRF}
   & Scaffold-GS & 174.0 & \seagg{0.197}  & \seagg{0.219}  & 132.69          & \seagg{0.952}  & \seaggg{24.75} & \seaggg{0.840} \\
   & +WD         & 164.5 & 0.218          & 0.228          & \seagg{118.53}  & 1.028          & 24.02          & 0.788          \\
   & +WD-R       & 169.2 & \seaggg{0.193} & \seaggg{0.207} & \seaggg{112.66} & \seaggg{0.902} & \seagg{24.58}  & \seagg{0.829}  \\
  \bottomrule
  \end{tabular}%
  }
\end{table}

\section{Additional visual comparisons}

\subsection{3DGS representation}
\label{appendix_sec:representation_visualization}

\begin{figure*}[h]
\centering
\begin{tikzpicture}[
    x=.25\linewidth,
    y=-.19\linewidth,
    every node/.style={font=\footnotesize},
]
\node[right, inner sep=0pt] (preview) at (0, 0)
  {\includegraphics[width=.24\linewidth]{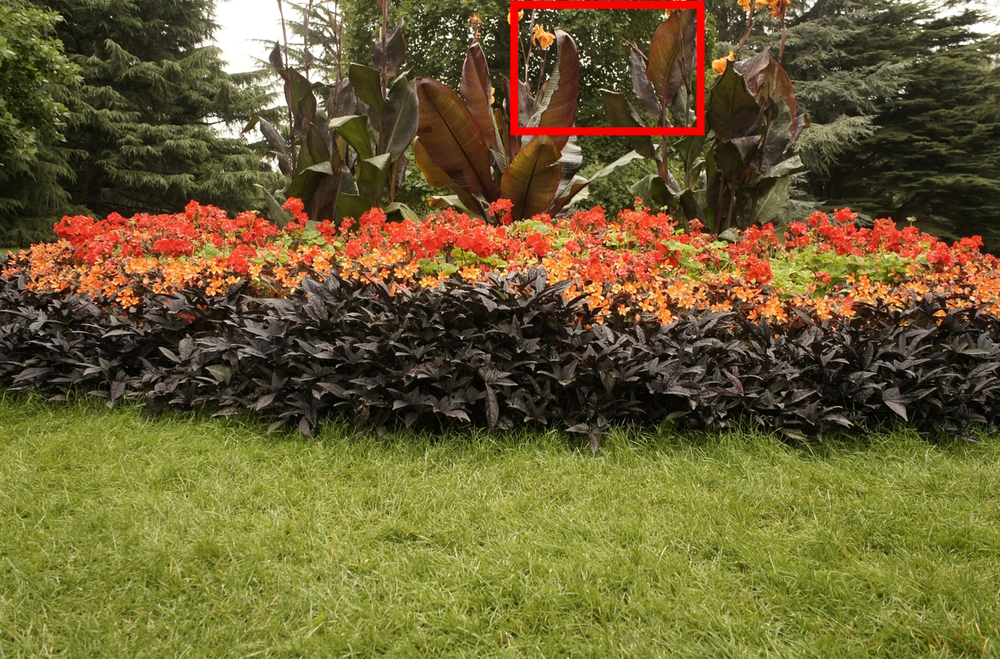}};
\node[left, inner sep=0pt] (gt) at (2, 0)
  {\includegraphics[width=.24\linewidth]{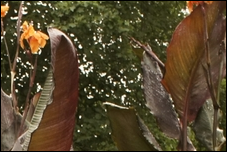}};
\node[left, inner sep=0pt] (orig) at (3, 0)
  {\includegraphics[width=.24\linewidth]{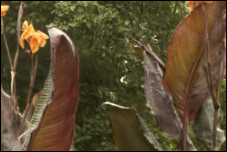}};
\node[left, inner sep=0pt] (wd) at (4, 0)
  {\includegraphics[width=.24\linewidth]{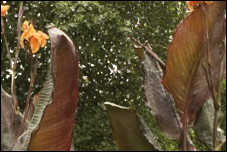}};
\node[left, inner sep=0pt] (pixelgs) at (2, 1.25)
  {\includegraphics[width=.24\linewidth]{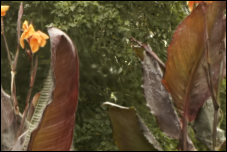}};
\node[left, inner sep=0pt] (perceptualgs) at (3, 1.25)
  {\includegraphics[width=.24\linewidth]{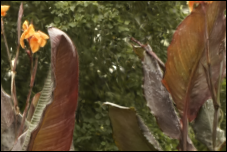}};
\node[left, inner sep=0pt] (wd_orig) at (4, 1.25)
  {\includegraphics[width=.24\linewidth]{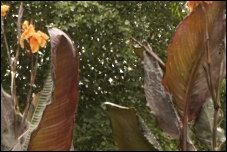}};

\node[below right=0 of preview.south west] {\texttt{Flowers}};
\node[below right=0 of gt.south west] {Ground truth};
    \node[below right=0 of orig.south west] {\shortstack[l]{Original loss~\cite{kerbl3Dgaussians} \\ (\#G: 3.38M)}};
\node[below right=0 of wd.south west] {\shortstack[l]{WD \\  (\#G: 2.80M)}};
\node[below right=0 of pixelgs.south west] {\shortstack[l]{Pixel-GS~\cite{zhang2024pixel} \\ (\#G: 7.08M)}};
\node[below right=0 of perceptualgs.south west] {\shortstack[l]{Perceptual-GS~\cite{zhou2025perceptual} \\ (\#G: 3.55M)}};
\node[below right=0 of wd_orig.south west] {\shortstack[l]{WD-R \\ (\#G: 2.73M)}};
\end{tikzpicture}
\begin{tikzpicture}[
    x=.25\linewidth,
    y=-.19\linewidth,
    every node/.style={font=\small},
]
\node[right, inner sep=0pt] (preview) at (0, 0)
  {\includegraphics[width=.24\linewidth]{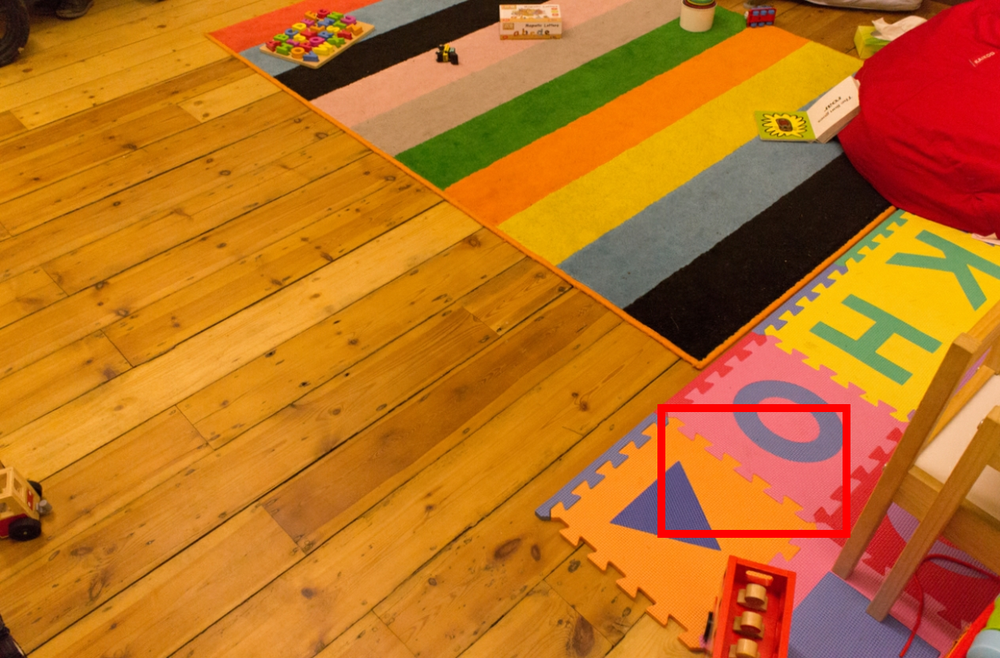}};
\node[left, inner sep=0pt] (gt) at (2, 0)
  {\includegraphics[width=.24\linewidth]{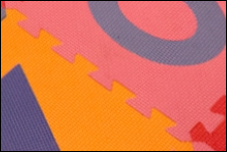}};
\node[left, inner sep=0pt] (orig) at (3, 0)
  {\includegraphics[width=.24\linewidth]{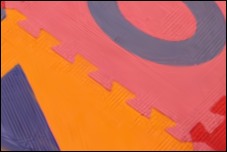}};
\node[left, inner sep=0pt] (wd) at (4, 0)
  {\includegraphics[width=.24\linewidth]{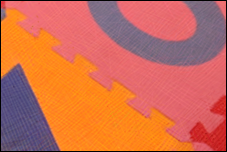}};
\node[left, inner sep=0pt] (pixelgs) at (2, 1.25)
  {\includegraphics[width=.24\linewidth]{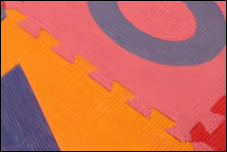}};
\node[left, inner sep=0pt] (perceptualgs) at (3, 1.25)
  {\includegraphics[width=.24\linewidth]{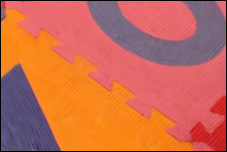}};
\node[left, inner sep=0pt] (wd_orig) at (4, 1.25)
  {\includegraphics[width=.24\linewidth]{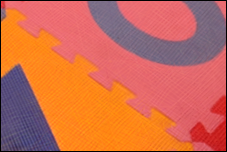}};
\node[below right=0 of preview.south west] {\texttt{Playroom}};
\node[below right=0 of gt.south west] {Ground truth};
\node[below right=0 of orig.south west] {\shortstack[l]{Original loss\\ (\#G: 2.33M)}};
\node[below right=0 of wd.south west] {\shortstack[l]{WD \\ (\#G: 2.35M)}};
\node[below right=0 of pixelgs.south west] {\shortstack[l]{Pixel-GS \\ (\#G: 3.76M)}};
\node[below right=0 of perceptualgs.south west] {\shortstack[l]{Perceptual-GS \\ (\#G: 2.29M)}};
\node[below right=0 of wd_orig.south west] {\shortstack[l]{WD-R \\ (\#G: 2.31M)}};
\end{tikzpicture}
\begin{tikzpicture}[
    x=.25\linewidth,
    y=-.19\linewidth,
    every node/.style={font=\small},
]
\node[right, inner sep=0pt] (preview) at (0, 0)
  {\includegraphics[width=.24\linewidth]{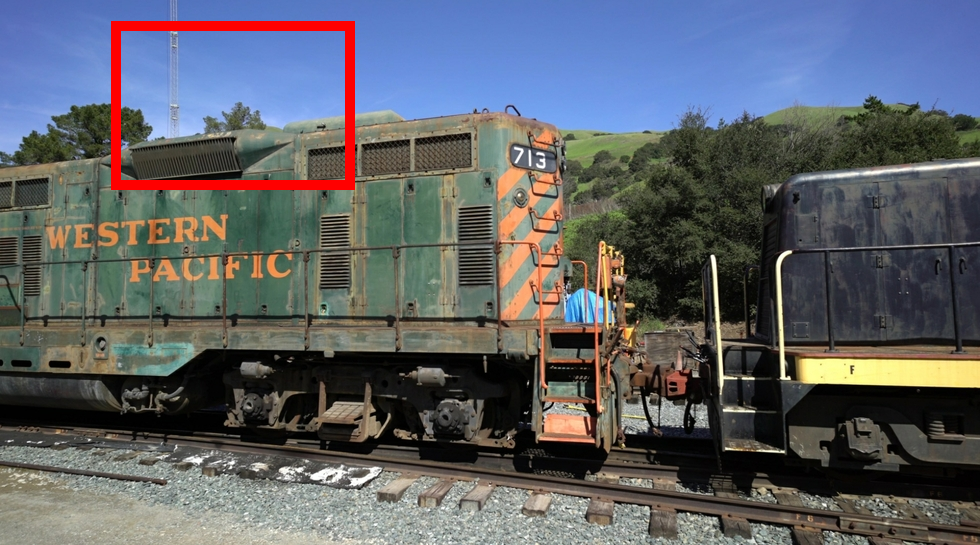}};
\node[left, inner sep=0pt] (gt) at (2, 0)
  {\includegraphics[width=.24\linewidth]{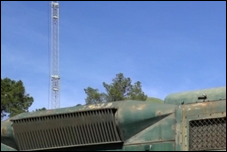}};
\node[left, inner sep=0pt] (orig) at (3, 0)
  {\includegraphics[width=.24\linewidth]{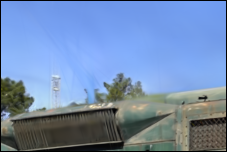}};
\node[left, inner sep=0pt] (wd) at (4, 0)
  {\includegraphics[width=.24\linewidth]{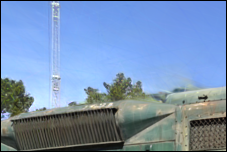}};
\node[left, inner sep=0pt] (pixelgs) at (2, 1.25)
  {\includegraphics[width=.24\linewidth]{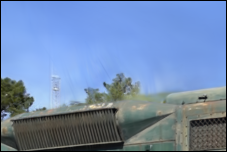}};
\node[left, inner sep=0pt] (perceptualgs) at (3, 1.25)
  {\includegraphics[width=.24\linewidth]{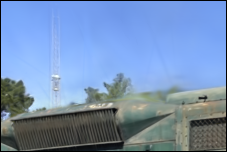}};
\node[left, inner sep=0pt] (wd_orig) at (4, 1.25)
  {\includegraphics[width=.24\linewidth]{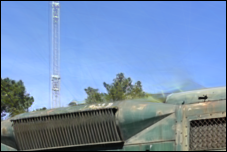}};
\node[below right=0 of preview.south west] {\texttt{Train}};
\node[below right=0 of gt.south west] {Ground truth};
\node[below right=0 of orig.south west] {\shortstack[l]{Original loss \\ (\#G: 1.08M)}};
\node[below right=0 of wd.south west] {\shortstack[l]{WD \\ (\#G: 1.50M)}};
\node[below right=0 of pixelgs.south west] {\shortstack[l]{Pixel-GS\\ (\#G: 3.80M)}};
\node[below right=0 of perceptualgs.south west] {\shortstack[l]{Perceptual-GS \\ (\#G: 1.39M)}};
\node[below right=0 of wd_orig.south west] {\shortstack[l]{WD-R \\ (\#G: 1.45M)}};
\end{tikzpicture}
\caption{Visual comparison of the novel view synthesis results obtained by the original 3DGS~\cite{kerbl3Dgaussians}, Pixel-GS~\cite{zhang2024pixel}, Perceptual-GS~\cite{zhou2025perceptual}, and the perceptual loss families discussed in~\cref{subsec:perceptual_losses}. The left images show the full scenes, with detailed crops highlighting reconstruction differences across methods, where $\#\mathrm{G}$ indicates the number of Gaussian splats for each method.}
\label{fig:generation_comparison_appendix}
\end{figure*}

For further visual comparisons, see \cref{fig:generation_comparison_appendix}. WD and WD-R generally lead to better reproduction of texture and fine visual detail, at a comparable or better splat count.

\subsection{3DGS scene compression}
\label{appendix_subsec:compression_visualization}

\begin{figure*}[htbp]
    \centering
\begin{tikzpicture}[
    x=.333\linewidth,
    y=-.25\linewidth,
    every node/.style={font=\small},
]
\node[right, inner sep=0pt] (preview) at (0, 0)
  {\includegraphics[width=.32\linewidth]{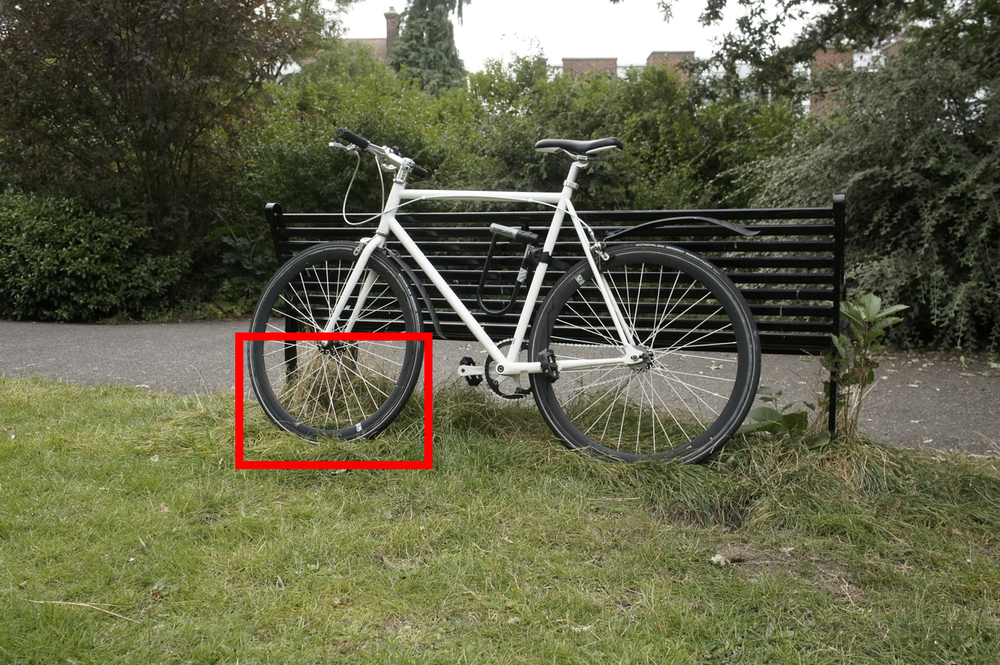}};
\node[left, inner sep=0pt] (gt) at (2, 0)
  {\includegraphics[width=.32\linewidth]{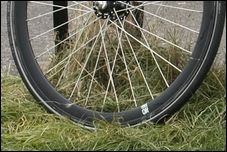}};
\node[left, inner sep=0pt] (orig) at (3, 0)
  {\includegraphics[width=.32\linewidth]{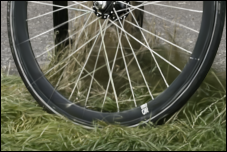}};
\node[left, inner sep=0pt] (composite) at (1, 1)
  {\includegraphics[width=.32\linewidth]{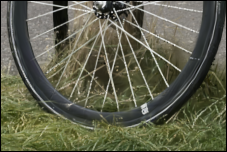}};
\node[left, inner sep=0pt] (wd) at (2, 1)
  {\includegraphics[width=.32\linewidth]{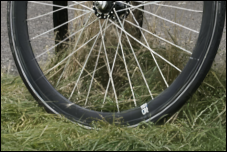}};
\node[left, inner sep=0pt] (wd_orig) at (3, 1)
  {\includegraphics[width=.32\linewidth]{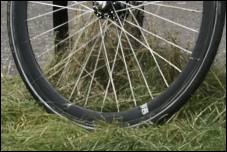}};
\node[below right=0 of preview.south west] {\texttt{Bicycle}};
\node[below right=0 of gt.south west] {Ground truth};
\node[below right=0 of orig.south west] {Original loss~\cite{kerbl3Dgaussians} (26.16 MB)};
\node[below right=0 of composite.south west] {Composite (25.01 MB)};
\node[below right=0 of wd.south west] {WD (24.79 MB)};
\node[below right=0 of wd_orig.south west] {WD-R (24.68 MB)};
\end{tikzpicture}
\begin{tikzpicture}[
    x=.333\linewidth,
    y=-.25\linewidth,
    every node/.style={font=\small},
]
\node[right, inner sep=0pt] (preview) at (0, 0)
  {\includegraphics[width=.32\linewidth]{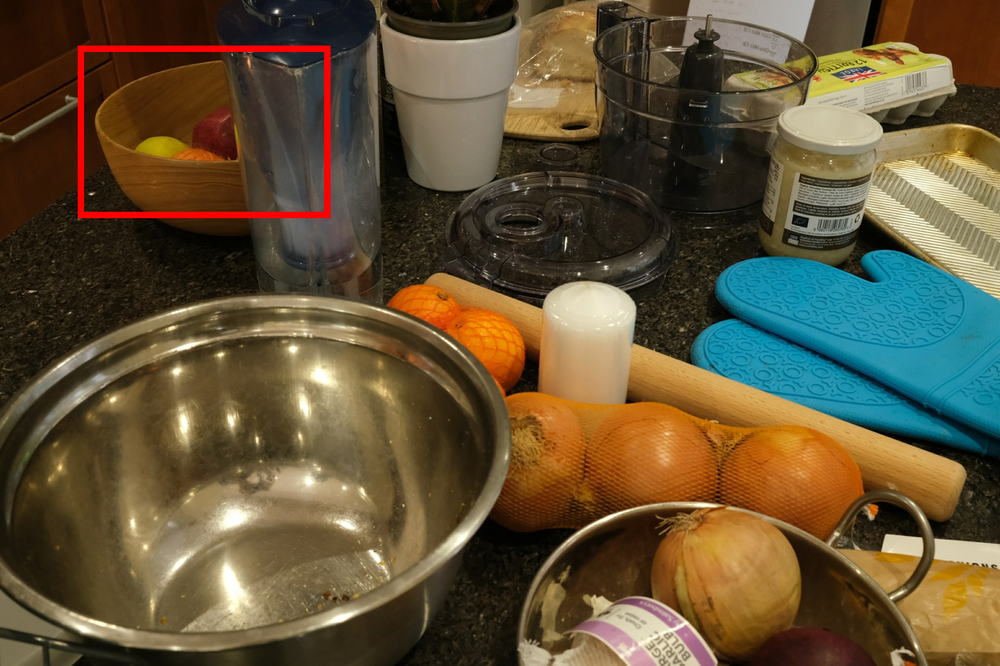}};
\node[left, inner sep=0pt] (gt) at (2, 0)
  {\includegraphics[width=.32\linewidth]{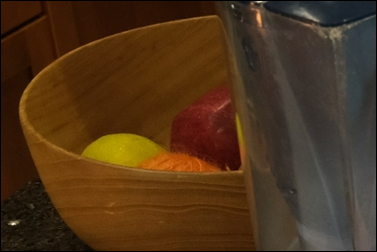}};
\node[left, inner sep=0pt] (orig) at (3, 0)
  {\includegraphics[width=.32\linewidth]{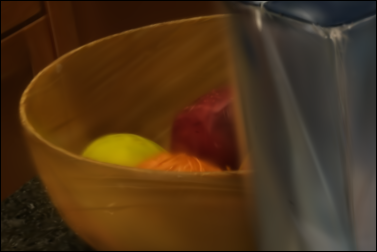}};
\node[left, inner sep=0pt] (composite) at (1, 1)
  {\includegraphics[width=.32\linewidth]{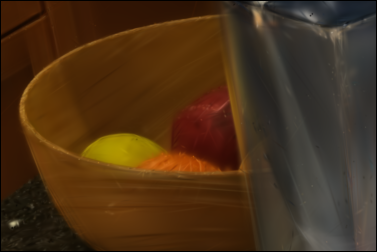}};
\node[left, inner sep=0pt] (wd) at (2, 1)
  {\includegraphics[width=.32\linewidth]{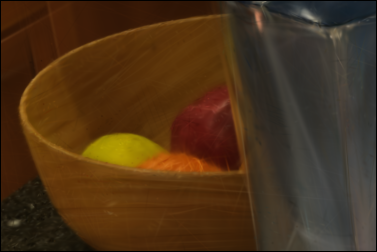}};
\node[left, inner sep=0pt] (wd_orig) at (3, 1)
  {\includegraphics[width=.32\linewidth]{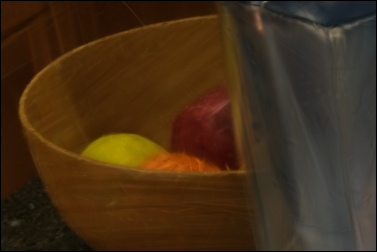}};
\node[below right=0 of preview.south west] {\texttt{Counter}};
\node[below right=0 of gt.south west] {Ground truth};
\node[below right=0 of orig.south west] {Original loss (11.11 MB)};
\node[below right=0 of composite.south west] {Composite (10.92 MB)};
\node[below right=0 of wd.south west] {WD (10.77 MB)};
\node[below right=0 of wd_orig.south west] {WD-R (11.13 MB)};
\end{tikzpicture}
\caption{Visual comparison of the novel view synthesis results obtained by the Comp-GS~\cite{liu2024compgs} compression algorithm. The left images show the full scenes, with detailed crops highlighting reconstruction differences across methods, where MB indicates the storage size for each method.}
\label{fig:compression_comparison_appendix}
\end{figure*}

Further visual comparisons are provided in~\cref{fig:compression_comparison_appendix}, where WD and WD-R consistently reproduce textures and fine visual details more faithfully at comparable storage sizes.

\section{Erank histograms per scene for BungeeNeRF dataset}
\label{appendix_sec:erank_details}
\begin{figure}[htbp]
\begin{subfigure}{0.5\textwidth}
\includegraphics[width=\textwidth]{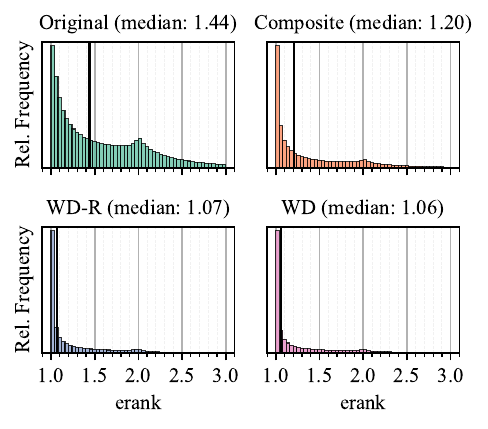}
\caption{\texttt{Amsterdam}}
\end{subfigure}%
\hfill%
\begin{subfigure}{0.5\textwidth}
\includegraphics[width=\textwidth]{figures/erank_hist/barcelona_density_histogram.pdf}
\caption{\texttt{Barcelona}}
\end{subfigure}%
\\%
\begin{subfigure}{0.5\textwidth}
\includegraphics[width=\textwidth]{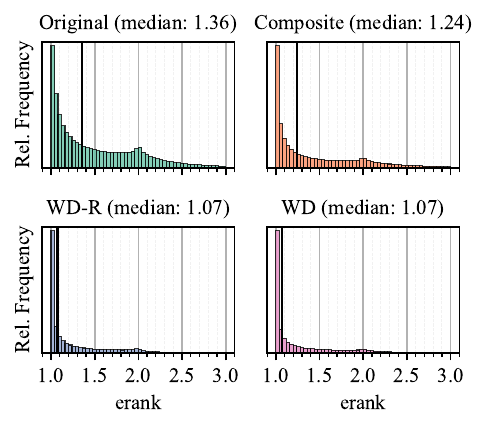}
\caption{\texttt{Bilbao}}
\end{subfigure}%
\hfill%
\begin{subfigure}{0.5\textwidth}
\includegraphics[width=\textwidth]{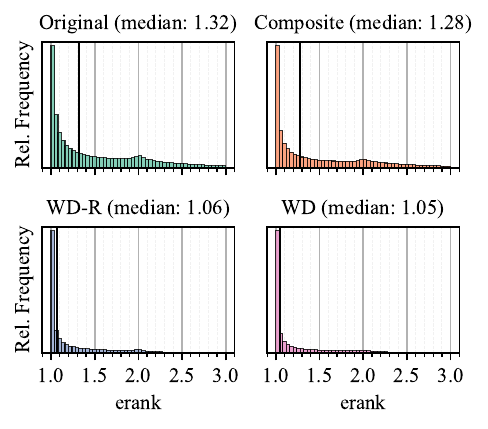}
\caption{\texttt{Chicago}}
\end{subfigure}
\caption{Effective rank (erank)~\cite{hyung2024effective} statistics for the \texttt{Amsterdam}, \texttt{Barcelona}, \texttt{Bilbao} and \texttt{Chicago} scenes in the BungeeNeRF dataset~\cite{xiangli2022bungeenerf}. The histograms show relative frequencies of the erank of all Gaussian covariance matrices in the scene, normalized to the maximum bin value. A diminished tail indicates that a larger fraction of the erank values is concentrated in the leftmost bin. This distribution shift is also reflected by the location of the median (black line).}
\label{fig:erank_histograms_part1}
\end{figure}

\begin{figure}[htbp]
\begin{subfigure}{0.5\textwidth}
\includegraphics[width=\textwidth]{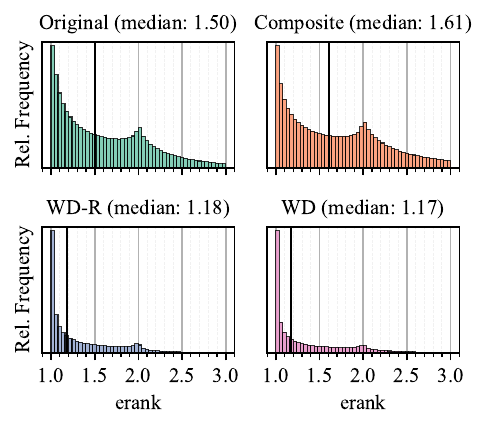}
\caption{\texttt{Hollywood}}
\end{subfigure}%
\hfill%
\begin{subfigure}{0.5\textwidth}
\includegraphics[width=\textwidth]{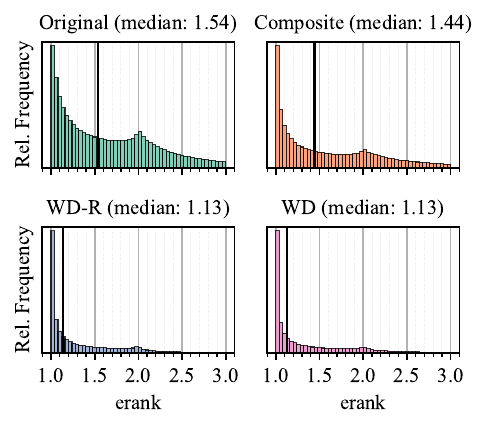}
\caption{\texttt{Pompidou}}
\end{subfigure}%
\\%
\begin{subfigure}{0.5\textwidth}
\includegraphics[width=\textwidth]{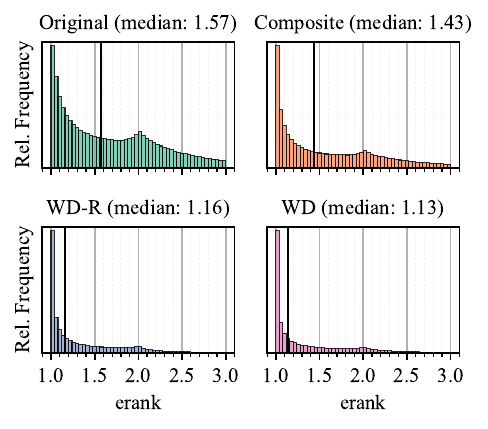}
\caption{\texttt{Quebec}}
\end{subfigure}%
\hfill%
\begin{subfigure}{0.5\textwidth}
\includegraphics[width=\textwidth]{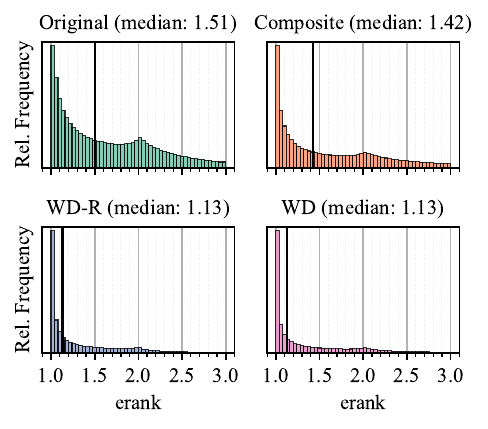}
\caption{\texttt{Rome}}
\end{subfigure}
\caption{Effective rank (erank)~\cite{hyung2024effective} statistics for the \texttt{Hollywood}, \texttt{Pompidou}, \texttt{Quebec} and \texttt{Rome} scenes in the BungeeNeRF dataset~\cite{xiangli2022bungeenerf}. The histograms show relative frequencies of the erank of all Gaussian covariance matrices in the scene, normalized to the maximum bin value. A diminished tail indicates that a larger fraction of the erank values is concentrated in the leftmost bin. This distribution shift is also reflected by the location of the median (black line).}
\label{fig:erank_histograms_part2}
\end{figure}

Additional effective rank (\emph{``erank''}) statistics for each scene of the BungeeNeRF dataset~\cite{xiangli2022bungeenerf} are given in~\cref{fig:erank_histograms_part1,fig:erank_histograms_part2}. The data corroborates the results from~\cref{fig:erank_hist_barcelona} in the main text. The loss functions using WD generally lead to higher anisotropy of the fitted Gaussians.


\section{Splat density heatmaps}
\label{appendix_sec:density_heatmaps}

To complement the \emph{erank} analysis (\cref{appendix_sec:erank_details}), which characterizes how perceptual optimization shapes individual Gaussians, we also visualize how \mbox{WD-R} distributes splat capacity \emph{spatially} across a rendered view.

\paragraph{\textbf{Computation.}}
For a given trained model and target camera view, we project each Gaussian center $\boldsymbol{\mu}_i$ onto the image plane using the camera's intrinsic and extrinsic parameters; only Gaussians whose projected centers fall within the image are kept, and we refer to these as the \emph{visible splats} for that view. For each remaining splat $G_i$, we add its opacity $\alpha_i$ to the $2{\times}2$-pixel bin containing the projected center, yielding a 2D density map. Because the map is computed from splat centers---not from the alpha-blended renderer---it captures where representation capacity is allocated.

\begin{figure}[t]
    \centering
    \begin{subfigure}{0.75\linewidth}\centering
        \includegraphics[width=\linewidth]{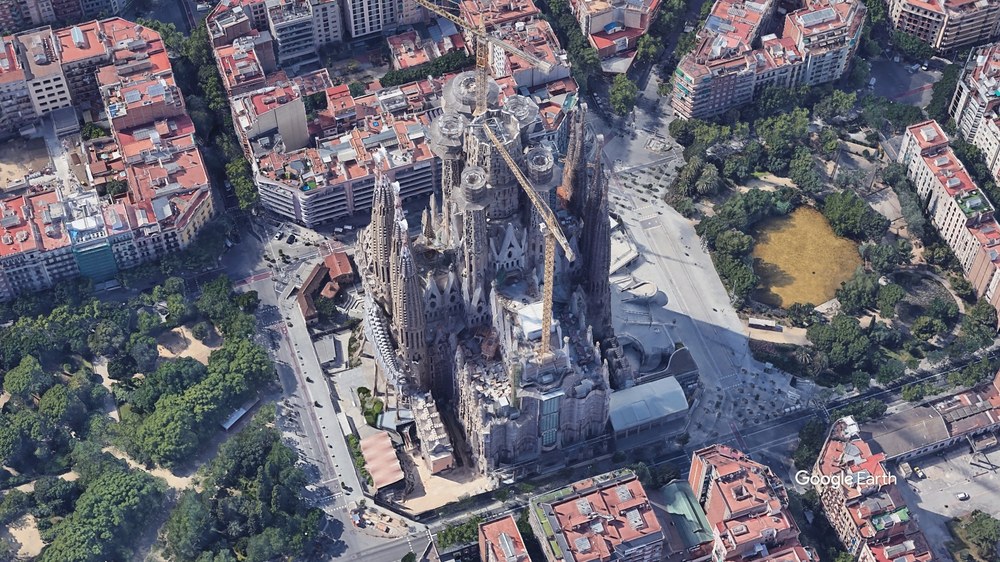}
        \caption{\texttt{Barcelona} view (ground truth).}
    \end{subfigure}\\[0.4em]
    \begin{subfigure}{0.75\linewidth}\centering
        \includegraphics[width=\linewidth]{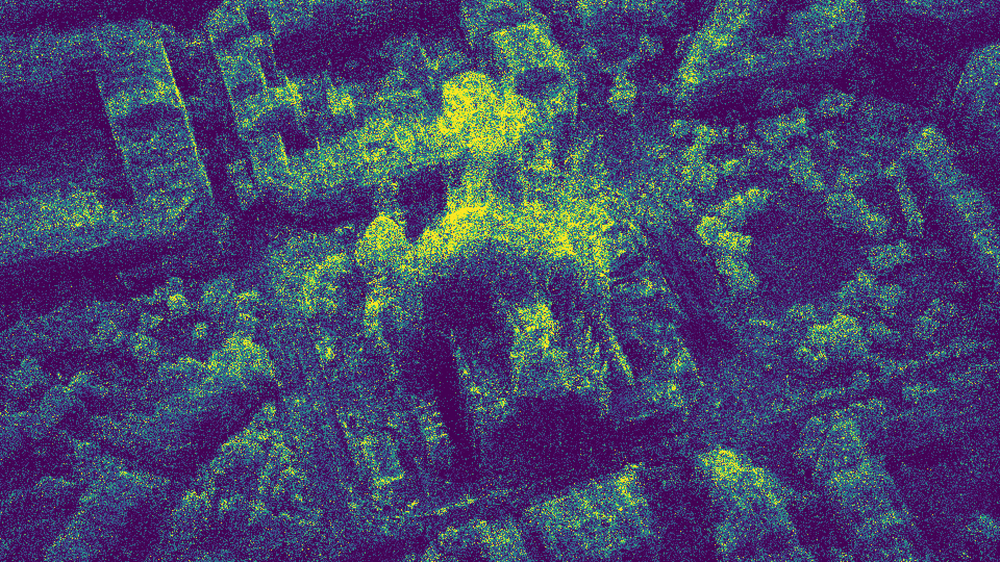}
        \caption{Original 3DGS~\cite{kerbl3Dgaussians}.}
    \end{subfigure}\\[0.4em]
    \begin{subfigure}{0.75\linewidth}\centering
        \includegraphics[width=\linewidth]{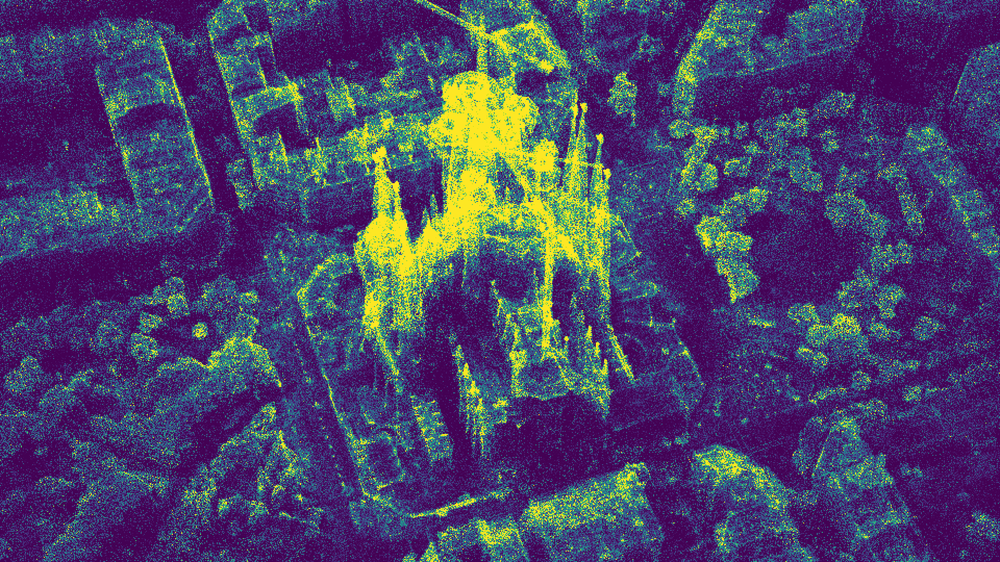}
        \caption{\mbox{WD-R} (ours).}
    \end{subfigure}
    \caption{Splat density heatmaps for one view of the \texttt{Barcelona} scene from BungeeNeRF~\cite{xiangli2022bungeenerf}. Each heatmap (b, c) shows the cumulative opacity of projected splat centers per $2{\times}2$-pixel bin. \mbox{WD-R} concentrates representation capacity on the texture-rich cathedral, while original 3DGS~\cite{kerbl3Dgaussians} spreads splats more uniformly across the view. Aggregated across views, \mbox{WD-R} uses fewer total splats than original 3DGS but exposes ${\sim}10\%$ more visible splats per view, suggesting more efficient view-relevant allocation.}
    \label{fig:density_heatmap_appendix}
\end{figure}

\paragraph{\textbf{Observations.}}
\cref{fig:density_heatmap_appendix} provides a larger version of \cref{fig:density_heatmap_main} from the main text, showing splat density on one view of the \texttt{Barcelona} scene from the BungeeNeRF dataset~\cite{xiangli2022bungeenerf}, comparing original 3DGS~\cite{kerbl3Dgaussians} and our \mbox{WD-R} model trained at a comparable Gaussian budget. Original 3DGS spreads splats more uniformly---including over low-detail ground regions---while \mbox{WD-R} concentrates capacity on the texture-rich cathedral. Aggregated across views, \mbox{WD-R} uses fewer total splats than original 3DGS but exposes ${\sim}10\%$ more visible splats per view, suggesting more efficient view-relevant allocation. Note that, similar to the erank analysis in \cref{subsec:3dgs_representation}, this is a post-hoc observation rather than a causal explanation of the perceptual gains.

\end{document}